\documentclass[10pt,journal,compsoc]{IEEEtran}
\ifCLASSOPTIONcompsoc
  \usepackage[nocompress]{cite}
\else
  \usepackage{cite}
\fi
\usepackage[hyphens]{url}
\usepackage{hyperref}
\hypersetup{colorlinks=true,breaklinks=true}
\usepackage{dsfont}
\usepackage{amsmath}
\usepackage{float}
\usepackage{mathtools}
\usepackage[T1]{fontenc}
\usepackage{amsmath}
\usepackage{caption}
\usepackage{subcaption}
\usepackage{textalpha}
\usepackage{dsfont}
\usepackage{amsmath}
\usepackage{float}
\usepackage{multirow}
\usepackage{hhline}
\usepackage{longtable}
\usepackage{caption}
\usepackage{xcolor,pifont}
\usepackage{multicol}
\usepackage[left=17.5mm,right=17.5mm,top=24.5mm,bottom=33.95mm]{geometry}
\usepackage{tabularx}
\usepackage{amssymb}
\usepackage{pifont}
\usepackage{adjustbox}
\usepackage{lipsum}
\usepackage{booktabs}
\usepackage{xcolor}

\usepackage{xcolor,pifont}
\newcommand*\colourcheck[1]{%
  \expandafter\newcommand\csname #1check\endcsname{\textcolor{#1}{\ding{52}}}%
}
\colourcheck{blue}
\colourcheck{green}
\colourcheck{red}
\usepackage{multirow}
\usepackage{hhline}

\ifCLASSINFOpdf
\else
\fi
\begin{document}
%
\title{Object Detection with Transformers: A Review}
%
%
%
%

\author{Tahira~Shehzadi,
        Khurram~Azeem~Hashmi,
        Didier~Stricker
        and~Muhammad~Zeshan~Afzal
\IEEEcompsocitemizethanks{\IEEEcompsocthanksitem All the members are with Department of Computer Science Technical University of Kaiserslautern, Mindgarage Lab, 
German Research Institute for Artificial Intelligence (DFKI)
Kaiserslautern, Germany 67663\protect\\
E-mail: Tahira.Shehzadi@dfki.de 
}
}

\IEEEtitleabstractindextext{%
\begin{abstract}
The astounding performance of transformers in natural language processing (NLP) has motivated researchers to explore their applications in computer vision tasks. DEtection TRansformer (DETR) introduces transformers to object detection tasks by reframing detection as a set prediction problem. Consequently, eliminating the need for proposal generation and post-processing steps. Initially, despite competitive performance, DETR suffered from slow training convergence and ineffective detection of smaller objects. However, numerous improvements are proposed to address these issues, leading to substantial improvements in DETR and enabling it to exhibit state-of-the-art performance. To our knowledge, this is the first paper to provide a comprehensive review of 21 recently proposed advancements in the original DETR model. We dive into both the foundational modules of DETR and its recent enhancements, such as modifications to the backbone structure, query design strategies, and refinements to attention mechanisms. Moreover, we conduct a comparative analysis across various detection transformers, evaluating their performance and network architectures. We hope that this study will ignite further interest among researchers in addressing the existing challenges and exploring the application of transformers in the object detection domain. Readers interested in the ongoing developments in detection transformers can refer to our website at \href{https://github.com/mindgarage-shan/transformer_object_detection_survey}{https://github.com/mindgarage-shan/transformer\_object\_detection\_survey}.

\end{abstract}

}

\maketitle

\IEEEdisplaynontitleabstractindextext

%
\IEEEpeerreviewmaketitle

\IEEEraisesectionheading{\section{Introduction}\label{sec:introduction}}
Object detection is one of the fundamental tasks in computer vision that involves locating and classifying objects within an image \cite{fasterrcnn3,fastrcnn5,yolov3,retinaNet68}. Over the years, convolutional neural networks (CNNs) have been the primary backbone for object detection models \cite{fasterrcnn3}. However, the recent success of transformers in natural language processing (NLP) has led researchers to explore their potential in computer vision as well \cite{ViT23}. The transformer architecture \cite{att75} has been shown to be effective in capturing long-range dependencies in sequential data \cite{att75}, making it an attractive candidate for object detection tasks.

In 2020, Carion et al. proposed a novel object detection framework called DEtection TRansformer (DETR) \cite {detr34}, which replaces the traditional region proposal-based methods with a fully end-to-end trainable architecture that uses a transformer encoder-decoder network. The DETR network shows promising results, outperforming conventional CNN-based object detectors \cite{fasterrcnn3,fastrcnn5,yolov3,retinaNet68} while also eliminating the need for hand-crafted components such as region proposal networks and post-processing steps such as non-maximum suppression (NMS) \cite{rcnn13}.
\begin{figure*}[h]
\centering
\includegraphics[width=1\textwidth]{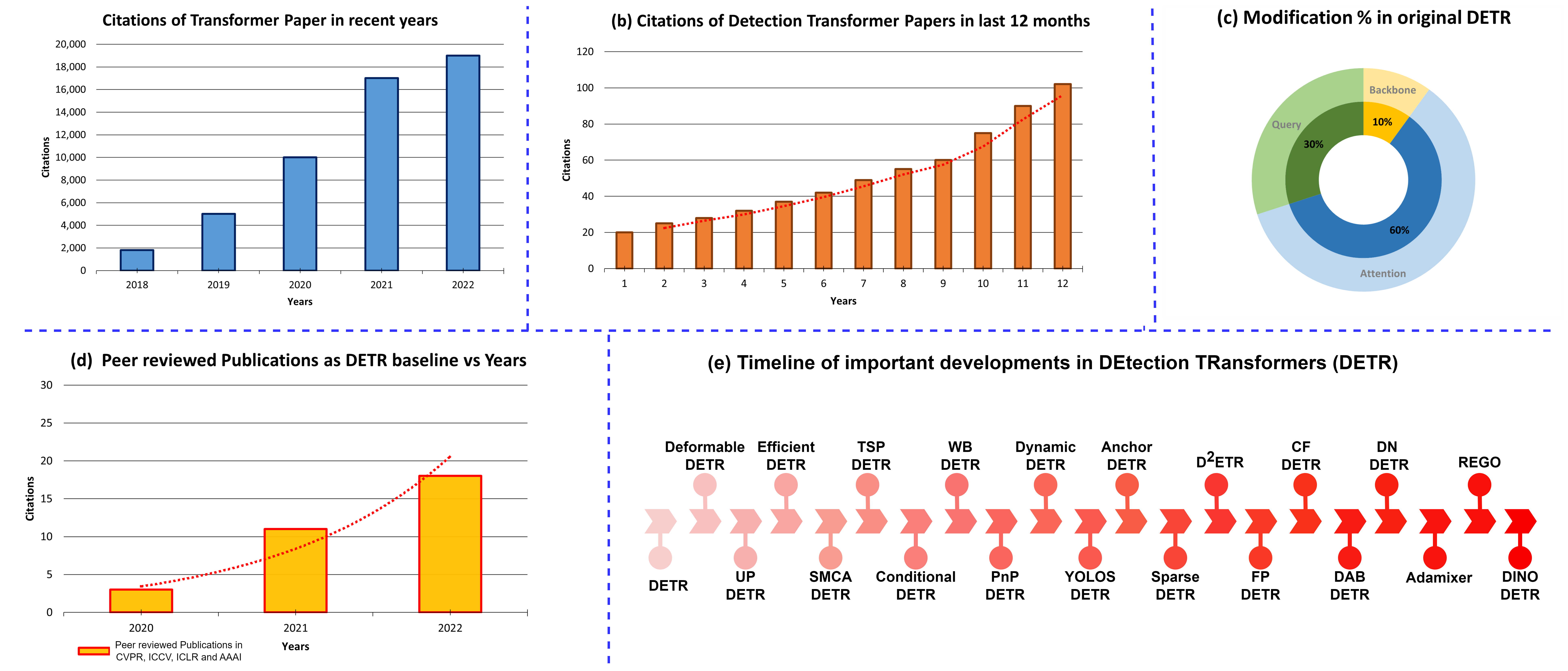}
\caption{Statistics overview of the literature on Transformers. (a) Number of citations per year of Transformer papers. (b) Citations in the last 12 months on Detection Transformer papers. (c) Modification percentage in original DEtection TRansformer (DETR) to improve performance and training convergence (d) Number of peer-reviewed publications per year that used DETR as a baseline. (e) A non-exhaustive timeline overview of important developments in DETR for detection tasks.}\label{fig:papers}
\end{figure*}

Since the introduction of DETR, several modifications and improvements have been proposed to overcome its limitations, such as slow training convergence and performance drops for small objects. Figure~\ref{fig:papers} shows the literature overview on the Detection Transformer and its modifications to improve performance and training convergence. Deformable-DETR \cite{Deformable54} modifies the attention modules to process the image feature maps by considering the attention mechanism as the main reason for slow training convergence. UP-DETR \cite{updetr23} proposes a few modifications to Pre-train the DETR similar to the pretraining of transformers in natural language processing. Efficient-DETR \cite{efficientDE} based on original DETR and Deformable-DETR examines the randomly initialized object probabilities, including reference points and object queries, which is one of the reasons for multiple training iterations. SMCA-DETR \cite{smca23} introduces a Spatially-Modulated Co-attention module that replaces the existing co-attention mechanism in DETR to overcome the slow training convergence of DETR. TSP-DETR \cite{Reth78} deals with the cross-attention and the instability of bipartite matching to overcome the slow training convergence of DETR. Conditional-DETR \cite{CondDE} presents a conditional cross-attention mechanism to solve the training convergence issue of DETR. WB-DETR \cite{WBdetr4} considers CNN backbone for feature extraction as an extra component and presents a transformer encoder-decoder network without a backbone. PnP-DETR \cite{pnp6} proposes a PnP sampling module to reduce spatial redundancy and make the transformer network computationally more efficient. Dynamic-DETR \cite{DynamicDE} introduces dynamic attention in the encoder-decoder network to improve training convergence. YOLOS-DETR \cite{yolos6} presents the transferability and versatility of the Transformer from image recognition to detection in the sequence aspect using the least information about the spatial design of the input and improves performance. Anchor-DETR \cite{Anchor-detr} proposes object queries as anchor points that are extensively used in CNN-based object detectors. Sparse-DETR \cite{sparsedetr} reduces the computational cost by filtering encoder tokens with learnable cross-attention maps. D$^2$ETR \cite{decoderonly60} uses the fine-fused feature maps in the decoder from the backbone network with a novel cross-scale attention module. FP-DETR \cite{fpdetr} reformulates the pretraining and fine-tuning stages for detection transformers. CF-DETR \cite{cfdetr} refines the predicted locations by utilizing local information, as incorrect bounding box location reduces performance on small objects. DN-DETR \cite{dn42} uses noised object queries as additional decoder input to reduce the instability of the bipartite-matching mechanism in DETR, which causes the slow convergence problem. AdaMixer \cite{adamixer7} considers the encoder an extra network between the backbone and decoder that limits the performance and slower the training convergence because of its design complexity. It proposes a 3D sampling process and a few other modifications in the decoder. REGO-DETR \cite{rego2} proposes an RoI-based method for detection refinement to improve the attention mechanism in the detection transformer. DINO \cite{dino23} considers positive and negative noised object queries to make training convergence faster and to enhance the performance on small objects. 

Due to the rapid progress of transformer-based detection methods, keeping track of new advancements is becoming increasingly challenging. Thus, a review of ongoing progress is necessary and would be helpful for the researchers in the field. This paper provides a detailed overview of recent advancements in detection transformers. Table~\ref{tab:overview} shows the overview of Detection Transformer (DETR) modifications to improve performance and training convergence. 

\begin{table*}[h!]
\tiny
\setlength\tabcolsep{0pt}
\setlength\extrarowheight{1pt}
\begin{center}
\caption{Overview of improvements in DEtection Transformer (DETR) to make training convergence faster and improve performance for small objects. Here, Bk represents the backbone, Pre denotes Pre-training, Attn indicates Attention, and Qry represents Query of the transformer network. A description of the main contributions is shown here.}\label{tab:overview}
\renewcommand{\arraystretch}{0.6} 
\begin{tabular*}{\textwidth}{@{\extracolsep{\fill}}lp{0.5cm}p{0.5cm}p{0.5cm}p{0.5cm}cc@{\extracolsep{\fill}}}
\toprule
\multirow{2}{*}{\textbf{Methods}} & \multicolumn{4}{c}{\textbf{Modifications}} & \multirow{2}{*}{\textbf{Publication}}& \multirow{2}{*}{\textbf{Highlights}}\\
\cline{2-5}
 & \textbf{Bk} & \textbf{Pre} & \textbf{Attn} & \textbf{Qry} &  \\
\toprule
\textbf{DETR} \cite{detr34}& - & - &-& - & ECCV 2020 & Transformer, Set-based prediction,  bipartite matching  \\
\toprule
Deformable-DETR \cite{Deformable54} &&& \greencheck && ICLR 2021 & Deformable-attention module\\
\midrule
UP-DETR \cite{updetr23} & & \greencheck &&& CVPR 2021 & Unsupervised pre-training, random query patch detection  \\
\midrule
Efficient-DETR \cite{efficientDE} &&&&\greencheck& arXiv 2021 & Refence point and top-k queries selection module\\
\midrule
SMCA-DETR \cite{smca23} &&& \greencheck && ICCV 2021  & Spatially-Modulated Co-attention module \\
\midrule
TSP-DETR \cite{tspdetr81} &&&\greencheck&& ICCV 2021 & TSP-FCOS and TSP-RCNN modules for cross attention \\
\midrule
Conditional-DETR \cite{CondDE} &&&&\greencheck& ICCV 2021 & Conditional spatial queries  \\
\midrule
WB-DETR \cite{WBdetr4} & \greencheck &&&&  ICCV 2021 & Encoder-decoder network without a backbone, LIE-T2T encoder module \\
\midrule
PnP-DETR \cite{pnp6} &&&\greencheck&& ICCV 2021 & PnP sampling module including pool sampler and poll sampler \\
\midrule
Dynamc-DETR \cite{DynamicDE}  &&&\greencheck&& ICCV 2021 & Dynamic attention in the encoder-decoder network \\
\midrule
YOLOS-DETR \cite{yolos6}  & & \greencheck &&& NeurIPS 2021 & Pretraining encoder network \\
\midrule
Anchor-DETR \cite{Anchor-detr} &&&\greencheck&\greencheck& AAAI 2022 & Row and Column decoupled-attention, object queries as anchor points \\
\midrule
Sparse-DETR \cite{sparsedetr} &&&\greencheck&& ICLR 2022 & Cross-attention map predictor, deformable-attention module \\
\midrule
$D^2$ETR \cite{decoderonly60} &&&\greencheck&& arXiv 2022 & Fine fused features, cross-scale attention module \\
\midrule
FP-DETR \cite{fpdetr} & \greencheck & \greencheck &&& ICLR 2022 & Multiscale tokenizer in place of CNN backbone, pretraining encoder network  \\
\midrule
CF-DETR \cite{cfdetr} &&&\greencheck&& AAAI 2022 & TEF module to capture spatial relationships,  a coarse and a fine layer in the decoder network\\
\midrule
DAB-DETR \cite{dab89} &&&&\greencheck& ICLR 2022  & Dynamic anchor boxes as object queries  \\
\midrule
DN-DETR \cite{dn42} &&&&\greencheck& CVPR 2022 & Positive noised object queries   \\
\midrule
AdaMixer \cite{adamixer7} &&&\greencheck&& CVPR 2022 & 3D sampling module, Adaptive mixing module in the decoder\\
\midrule
REGO \cite{rego2} &&&\greencheck&& CVPR 2022 & A multi-level recurrent mechanism and a
glimpse-based decoder \\
\midrule
DINO\cite{dino23} & &  &  & \greencheck&    arXiv 2022 & Contrastive denoising module, positive and negative noised object queries  \\
\bottomrule
\end{tabular*}
\end{center}
\end{table*} 
\subsection{Our Contributions}
\begin{enumerate}
\item \textbf{Detailed review of transformer-based detection methods from architectural perspective}. We categorize and summarize improvements in DEtection TRansformer (DETR) according to Backbone modifications, pre-training level, attention mechanism, query design, etc. The proposed analysis aims to help researchers to have a more in-depth understanding of the key components of detection transformers in terms of performance indicators.

\item \textbf{A performance evaluation of detection transformers.} We evaluate improvements in detection transformers using popular benchmark MS COCO \cite{coco14}. We also highlight the advantages and limitations of these approaches.

\item \textbf{Analysis of accuracy and computational complexity of improved versions of detection transformers.}
We present an evaluative comparison of state-of-the-art transformer-based detection methods w.r.t attention mechanism, backbone modification, and query design.
\item \textbf{Overview of key building blocks of detection transformers to improve performance further and future directions.}
We examine the impact of various key architectural design modules that impact network performance and training convergence to provide possible suggestions for future research.
\end{enumerate}
The remaining paper is arranged as follows. Section \ref{sec:surveys} discusses previous related surveys on transformers. Section \ref{sec:Tvision} is related to object detection and transformers in all types of vision. Section \ref{sec:modules} is the main part which explains the modifications in the detection transformers in detail. Section \ref{sec:datasetEV} is about evaluation protocol, and Section \ref{sec:comp} provides an evaluative comparison of detection transformers. Section \ref{sec:future} discusses open challenges and future directions. Finally, Section \ref{sec:conclusion} concludes the paper.
\begin{table*}[h!]
\tiny
\begin{center}
\caption{Overview of previous surveys on object detection. For each paper, the publication details are provided.}\label{tab:survey}
\renewcommand{\arraystretch}{0.7} 
\begin{tabular*}{\textwidth}{@{\extracolsep{\fill}}p{7.6cm}p{0.4cm}p{0.4cm}p{8.5cm}@{\extracolsep{\fill}}}
\toprule
\textbf{Title} & 
\textbf{Year} &
\textbf{Venue} & 
\textbf{Description}  \\
\toprule
Advanced Deep-Learning Techniques for Salient and
Category-Specific Object Detection: A Survey \cite{AdDL3} & 2018 & SPM & It overviews different domains of object detection, i.e. objectness detection (OD), salient OD and category-specific OD.\\
\midrule
Object Detection in 20 Years: A Survey \cite{OD20} & 2019 & TPAMI & This work gives an overview of the evolution of object detectors.\\
\midrule
Deep Learning for Generic Object Detection: A Survey \cite{GenOD2} & 2019 & IJCV & A review on deep learning techniques on generic object detection.\\
\midrule
A Survey on Deep Learning-based Architectures for Semantic Segmentation on 2D images \cite{segsurvey3} & 2020 & PRJ & Deep learning-based methods for Semantic Segmentation are reviewed.\\
\midrule
A Survey of Modern Deep Learning based Object Detection Models \cite{gh92}  & 2021 & ICV & It briefly overviews deep learning-based (regression-based single-stage and candidate-based two-stage) object detectors.\\
\midrule
A Survey of Object Detection Based on CNN and Transformer \cite{deeptran5} & 2021 & PRML & A review of the benefits and drawbacks of deep learning-based object detectors and introduction of transformer-based methods. \\
\midrule
Transformers in computational visual media: A survey \cite{Tsvision3} & 2021 & CVM & It focuses on backbone design and low-level vision using vision transformer methods.\\
\midrule
A survey: object detection methods from CNN to transformer \cite{ark22survey}& 2022 & MTA & Comparison of various CNN-based detection networks and introduction of Transformer-based detection networks.\\
\midrule
A Survey on Vision Transformer \cite{VTsurvey6}& 2023 & TPAMI & This paper provides an overview of vision transformers and focuses on summarizing the state-of-the-art research in the field of Vision Transformers (ViTs).\\
\bottomrule
\end{tabular*}
\end{center}
\end{table*} 
\section{Related Previous Reviews and Surveys}
\label{sec:surveys}
Many surveys have studied deep learning approaches in object detection \cite{AdvDNN15, salientOD3, smsurvey4, ODsuvey78, facesurvey3, FineG56}. Table~\ref{tab:survey} lists existing object detection surveys. Among these surveys, many studies comprehensively review approaches that process different 2D data types\cite{tesxtDsurvey5, attenmodel4, AdDL3, GenOD2}. Other studies focus on specific 2D applications \cite{pedD6, segsurvey3, RsensingS7, vehicleD7,airportD5, trafficsurvey, Tsing5, tah9}  and other tasks such as segmentation \cite{sseg34,hyper34, insAware5}, image captioning \cite{DvSAll4, imcap5, imcap16, icap98} and object tracking \cite{ObjTrack4}. Furthermore, some surveys examine deep learning methods and introduce vision transformers \cite{ark22survey, deeptran5, Tsvision3, VTsurvey6}. However, most of the literature research was published before improvements in the detection transformer network, and a detailed review of transformer-based object detectors is missing. Thus, a survey of ongoing progress is necessary and would be helpful for researchers. 

\section{Object Detection and Transformers in Vision}
\label{sec:Tvision}
\subsection{Object Detection}
This section explains the key concept of object detection and previously used object detectors. A more detailed analysis of object detection concepts can be found in \cite{sv8,gh92,KAH54}. The object detection task localizes and recognizes objects in an image by providing a bounding box around each object and its category. These detectors are usually trained on datasets like PASCAL VOC \cite{pascalvoc3} or MS COCO \cite{coco14}. The backbone network extracts the features of the input image as feature maps \cite{fpn49}. Usually, the backbone network, such as the ResNet-50 \cite{resnet5}, is pre-trained on ImageNet \cite{NId6} and then finetuned to downstream tasks \cite{DETReg7, vfew45, semisup3, selfsup6, spatioKAH56, HashmiWACV}. Moreover, many works have also used visual transformers \cite{transback3, selsup5, YOLOv37} as a backbone. Single-stage object detectors \cite{ssd23,Joseph15, yolo9000, yolov3, yolov4, centernet2,retinaNet68,dssd7, Rssd4, refinedet, cornernet5} use only one  network having faster speed but lower performance than two-stage networks. 
Two-stage object detectors \cite{rcnn13, sppNet57, fastrcnn5, fasterrcnn3, fpn49, rfcn7, mask-rcnn84, DetectoRS5, ht8, cascadercnn8} contain two networks to provide final bounding boxes and class labels. 

\noindent\textbf{Lightweight Detectors:} Lightweight detectors are object detection models designed to be computationally efficient and require less computational resources than standard object detection models. These are real-time object detectors and can be employed on small devices. These networks include \cite{SqueezeNet6, MobileNets4, Mobilenetv2, mobilenetv3, ShuffleNet7, palee2, ShuffleNetv2, MnasNet5, OFA3}. 

\noindent\textbf{3D Object Detection:} The primary purpose of 3D object detection is to recognize the objects of interest using a 3D bounding box and give a class label. 3D approaches are divided into three categories as image-based \cite{Deepmanta7, 3dbboxG, GS3D5, Stereorcnn8, Geodd3, dle3d, YOLOStereo3D2}, point cloud-based \cite{cbdt5, VoxelNet6, PointPillar4, bhc5, CIASSD5, sessd3, voxelrccn2, imchW3, voxelTran5} and multimodal fusion-based \cite{PointPainting1, Joint3d4, Multisen4, sdcvf6, CLOCs4}.

\subsection{Transformer for Segmentation}
The self-attention mechanism can be employed for segmentation tasks \cite{seg15,seg101,seg134,seg135,seg81} that provides pixel-level \cite{seg93} prediction results. Panoptic segmentation \cite{seg136} jointly solves semantic and instance segmentation tasks by providing per-pixel class and instance labels. Wang et al.\cite{axial53} proposes location-sensitive axial attention for panoptic segmentation task on three benchmarks \cite{seg75,seg137,Cityscapes6}. The above segmentation approaches have self-attention in CNN-based networks. Recently segmentation transformers \cite{seg101,seg135} containing encoder-
decoder modules give new directions to employ transformers for segmentation tasks. 

\subsection{Transformers for Scene and Image Generation}
Previously, text-to-image generation methods \cite{igen158,igen159,igen160,igen161} are based on GANs \cite{gan61}. Ramesh et al. \cite{dallai7} introduced a transformer-based model for generating high-quality images from provided text details. Transformer networks are also applied for image synthesis \cite{igen23,igen143,igen144,igen145,igen146}, which is important for learning unsupervised and generative models for downstream tasks. The feature learning with an unsupervised training procedure \cite{igen143} achieves state-of-the-art performance on two datasets \cite{igen148,igen149}, while SimCLR \cite{igen150} provides comparable performance on \cite{seg138}. The iGPT mage generation network \cite{igen143} does not include pre-training procedures similar to language modeling tasks. However, unsupervised CNN-based networks \cite{igen117,igen151,igen152} consider prior knowledge as architectural layout, attention mechanism and regularization. Generative Adversarial Networks (GAN) \cite{gan61} with CNN-based backbones have been appealing for image synthesis \cite{rad155,igen156,style157}. TransGAN \cite{igen145} is a strong GAN network where the generator and discriminator contain transformer modules. These transformer-based networks boost performance for scene and image generation tasks.

\subsection{Transformers for Low-level Vision}
Low-level vision analyses images to identify their basic components and create an intermediate representation for further processing and higher-level tasks. After observing the remarkable performance of attention networks in high-level vision tasks \cite{detr34,seg101}, many attention-based approaches have been introduced for low-level vision problems, such as \cite{llv16,llv19,llv164,llv165,llv24}.
\begin{figure*}[h]
\centering
\includegraphics[width=13cm]{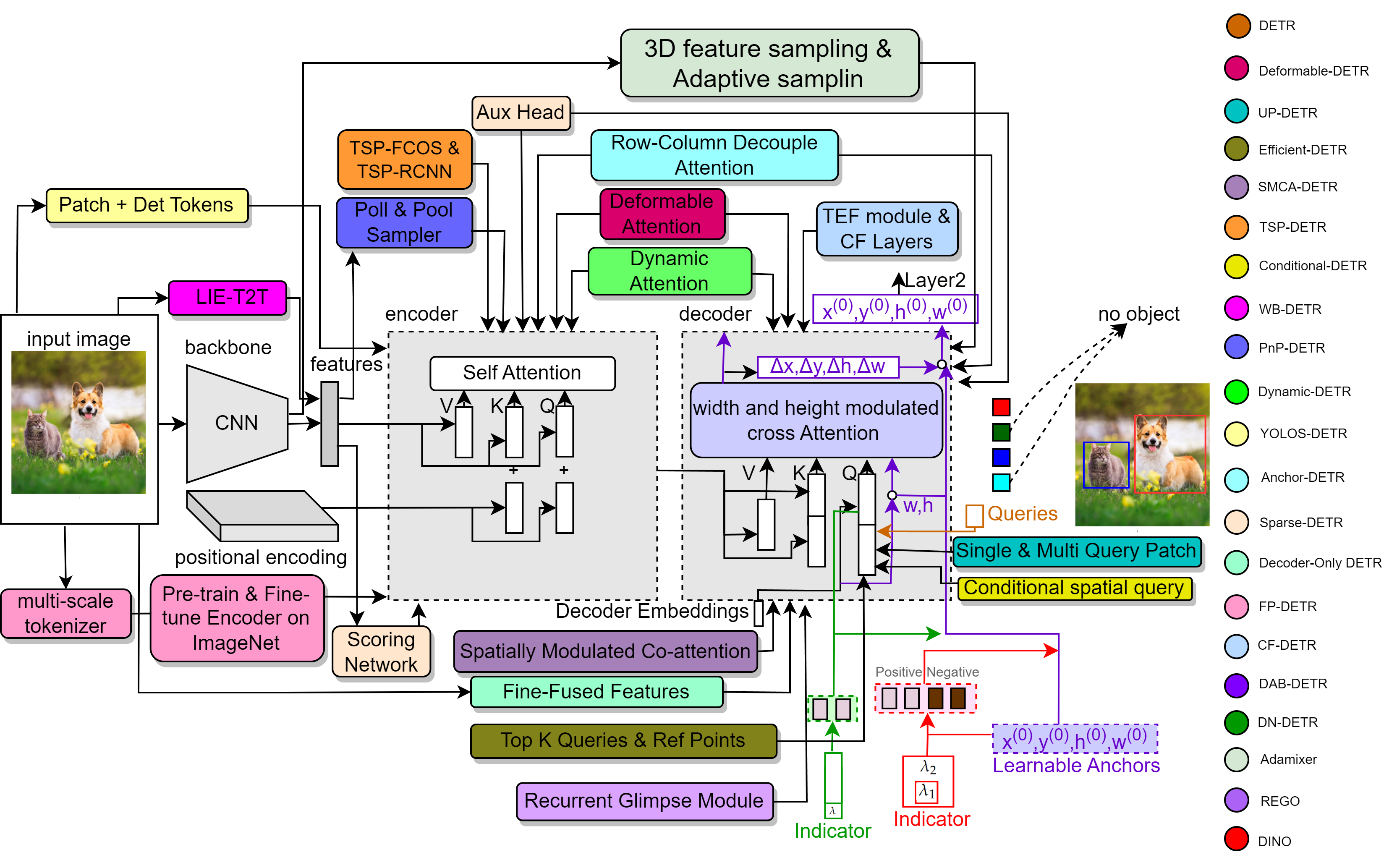}
\caption{An overview of the DEtection TRansformer (DETR) and its modifications proposed by recent methods to improve performance and training convergence. It considers the detection a set prediction task and uses the Transformer to free the network from post-processing steps such as non-maximal suppression (NMS). Here, each module added to DETR is represented with different color with its corresponding label (shown on the right side).}\label{fig:transformer}
\end{figure*}
\subsection{Transformers for Multi-Modal Tasks}
Multi-Modal Tasks involve processing and combining information from multiple sources or modalities, such as text, images, audio, or video. The application of transformer networks in vision language tasks has also been widespread, including visual question-answering \cite{mmt183}, visual commonsense-reasoning  \cite{mmt184}, cross-modal retrieval  \cite{mmt185}, and image captioning\cite{imgcap4}. These transformer designs can be classified into single-stream \cite{uniter43,oscar44,VideoBERT17,Unicoder180,VisualBERT63,VLBERT22} and dual-stream networks \cite{LXMERT21,ViLBERT181,vil182}. The primary distinction between these networks lies in the choice of loss functions.
\section{Detection Transformers}
\label{sec:modules}
This section briefly explains DEtection TRansformer (DETR) and its improvements as shown in Figure~\ref{fig:transformer}.
\subsection{DETR}
DEtection TRansformer (DETR) \cite{detr34} architecture is much simpler than CNN-based detectors like Faster R-CNN \cite{FasterTD} as it removes the need for anchors generation process and post-processing steps such as Non-Maximal Suppression (NMS) and provides an optimal detection framework.
The DETR network has three main modules: a backbone network with positional encodings, an encoder, and a decoder network with attention mechanism. The extracted features from the backbone network as one single vector and their positional encoding \cite{imgtrans45, attenaug} within the input vector fed to the encoder network. Here, the self-attention is performed on key, query, and value matrices forwarded to the multi-head attention and feed-forward network to find the attention probabilities of the input vector. The DETR decoder takes object queries in parallel with the encoder output. It computes predictions by decoding N number of object queries in parallel. It uses a bipartite-matching algorithm to label the ground-truth and predicted objects as given in the following equation: 
\setlength{\abovedisplayskip}{3pt}
\setlength{\belowdisplayskip}{3pt}
\begin{equation}
 \hat{\sigma} =\arg\min_{\sigma \in N} \sum_{k}^{N} \mathcal{L}_{m} (y_k,\hat{y}_{\sigma(k)}),
\end{equation}
Here, $y_k$ is a set of ground-truth (GT) objects. It provides boxes for both object and "no object" classes, where $N$ is the total number of objects to be detected. Here, $\mathcal{L}_{m} (y_k,\hat{y}_{\sigma(k)})$ is the matching cost (for direct prediction) without duplicates between predicted objects with index $\sigma(k)$ and ground-truth $y_k$ as shown in the following equation: 
\begin{equation}
\resizebox{.91\hsize}{!}{$\mathcal{L}_{m} (y_k,\hat{y}_{\sigma(k)})=-\mathds{1}_{\{c_k\neq \phi\}}\hat{p}_{\sigma(k)}(c_k)+\mathds{1}_{\{c_k\neq \phi\}}\mathcal{L}_{bbox}(b_k,\hat{b}_{\hat{\sigma}}(k))$}
\end{equation}
The next step is to compute the Hungarian loss by determining the optimal matching between ground-truth (GT) and detected boxes regarding bounding-box region and label. The loss is reduced by Stochastic Gradient Descent (SGD).
\begin{equation}
 \resizebox{.91\hsize}{!}{$\mathcal{L}_{H}(y, \hat y)= \sum_{k=1}^{N}[-log\hat{p}_{\hat{\sigma}(k)}(c_k)+\mathds{1}_{\{c_k\neq \phi\}}\mathcal{L}_{box}(b_k,\hat{b}_{\hat{\sigma}}(k))] $}
\end{equation}
Where $\hat{p}_{\hat{\sigma}(k)}$ and $c_k$ are the predicted class and target label, respectively. The term $\hat{\sigma}$ is the optimal-assignment factor, $b_k$ and $\hat{b}_{\hat{\sigma}}(k)$ are ground-truth and predicted bounding boxes. The term $\hat y$ and $y = \{(c_k, b_k)\}$ are the prediction and ground-truth of objects, respectively. Here, the bounding box loss is a linear combination of the generalized IoU (GIoU) loss \cite{GIOU2} and of the L1 loss, as in the following equation:
\begin{equation}
 \mathcal{L}_{bbox} = \lambda_{i} \mathcal{L}_{iou} (b_k,\hat{b}_{\sigma(k)}) + \lambda_{l1}\parallel b_k-\hat{b}_{\sigma(k)}\parallel_1
\end{equation}
Where $\lambda_{i}$ and $\lambda_{l1}$ are the hyperparameters. DETR can only predict a fixed number of $N$ objects in a single pass. For the COCO dataset \cite{coco14}, the value of $N$ is set to 100 as this dataset has 80 classes. This network doesn't need NMS to remove redundant predictions as it uses bipartite matching loss with parallel decoding \cite{WaveNet4, Auto81, maskpre4}. In comparison, previous works used RNNs-based autoregressive decoding \cite{pdcs15,RIS15,LDI4, LDI4, ISCR7, RnnSis6}. The DETR network has several challenges, such as slow training convergence and performance drops for small objects. To address these challenges, modifications have been made to the DETR network.
\begin{figure*}[h]
\centering
\includegraphics[width=14cm]{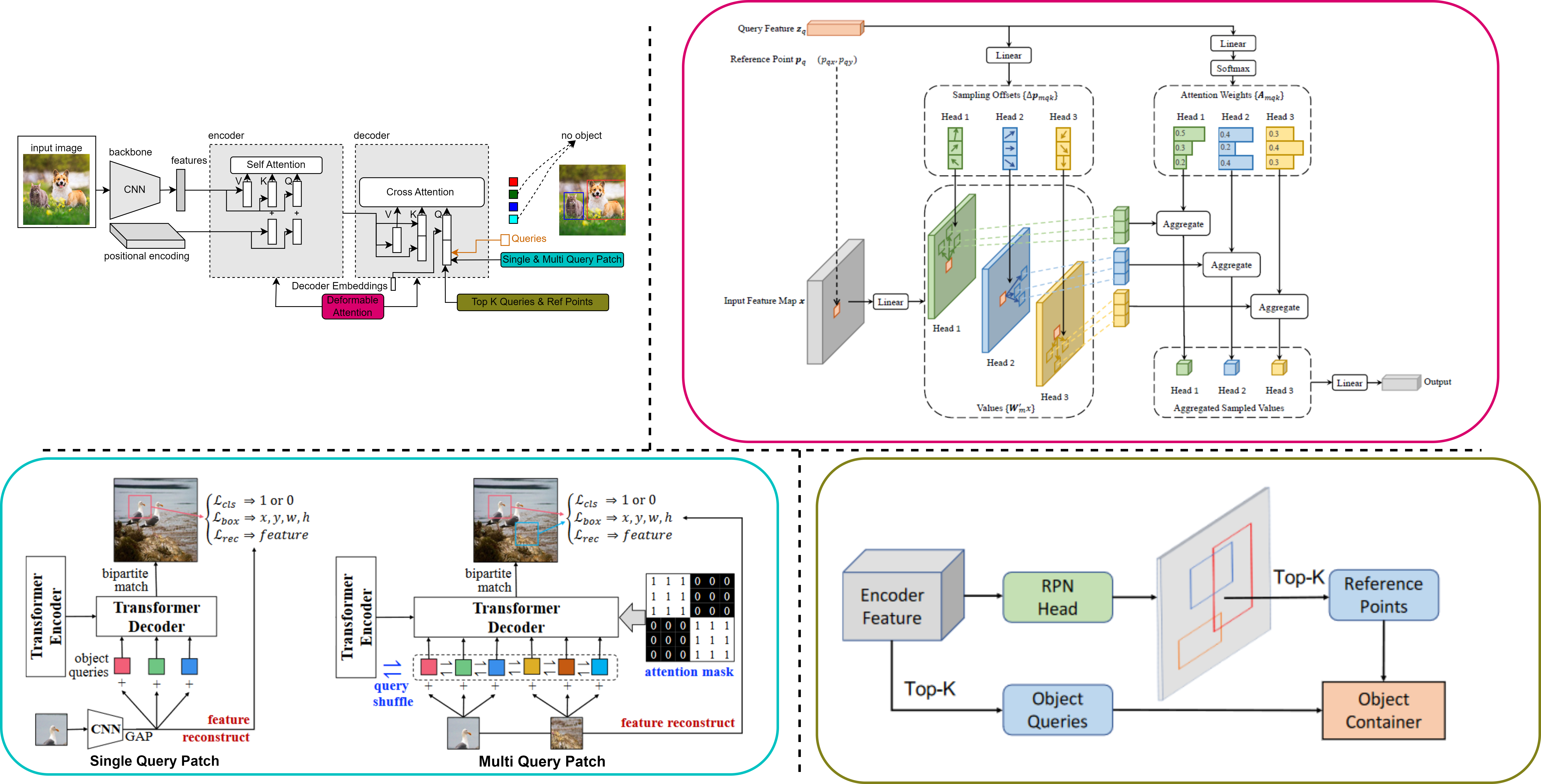}
\caption{The structure of the original DETR after the addition of Deformable-DETR \cite{Deformable54}, UP-DETR \cite{updetr23} and Efficient-DETR \cite{efficientDE}. Here, the top left network is a simple DETR network, along with improvement indicated with small colored boxes. Larger boxes with corresponding colored borders are utilized to illustrate the internal mechanisms of these small colored boxes. The top right block indicates Deformable-DETR, the bottom left block indicates UP-DETR, and the bottom right box represents Efficient-DETR (images from \cite{Deformable54, updetr23, efficientDE}).}\label{fig:deformable-up-efficient}
\end{figure*}
\subsection{Deformable-DETR}
The attention module of DETR  provides a uniform weight value to all pixels of the input feature map at the initialization stage. These weights need many epochs for training convergence to find informative pixel locations. However, it requires high computation and extensive memory. The computation complexity of self-attention in the encoder is $O (w_i^2h_i^2c_i)$, while the complexity of the cross-attention in the decoder is $O (h_i w_i c_i^2 + N h_iw_i c_i)$. Here, $h_i$ and $w_i$ denote the height and width of the input feature map, respectively, and N represents object queries fed as input to the decoder. Let $q\in \Omega _{q}$ denotes a query element with feature $z_q \in R^{c_i}$, and $k \in \Omega _k$ represents a key vector with feature $x_k \in R^{c_i}$, where $c_i$ is the input features dimension, $\Omega _{k}$ and $\Omega _{q}$ indicate the set of key and query vectors, respectively. Then, the feature of Multi-Head Attention (MHAttn) is computed by:
\begin{equation}
 \resizebox{.91\hsize}{!}{$ MHAttn(z_q, x)=\sum_{j=1}^{J}W_j[\sum_{k\in \Omega_{k}} A_{jqk}. W_j^\prime x_k] $}
\end{equation}
where j represents the attention head, ${W \prime}_j \in R^{c_v \times c_i}$ and $W_j \in R^{c_i\times c_v}$ are of learnable weights ($c_v = c_i/J$ by default). The attention weights $A_{jqk} \propto exp{\frac{ z_q^T U_j^T V_jx_k}{\sqrt c_v }}$ are normalized as $\sum_{k\in \Omega_k} A_{jqk} = 1$, in which $U_j, V_j \in R^{c_v\times c_i} $are also learnable weights. Deformable-DETR \cite{Deformable54} modifies the attention modules inspired by \cite{deformconv4, deformB3} to process the image feature map by considering the attention network as the main reason for slow training convergence and confined feature spatial resolution. This attention module works on taking a small number of samples nearby the reference point. Given an input feature map $x \in R^{c_i \times h_i \times w_i}$, let query q with content feature $z_q$ and a 2d reference point $r_q$, the deformable attention feature is computed by: 
\begin{equation}
 \resizebox{.91\hsize}{!}{$ DeformAttn(z_q, r_q,x)=\sum_{j=1}^{J}W_j[\sum_{k=1}^{K} A_{jqk}. W_j x(r_q + \Delta r_{jqk})] $}
\end{equation}
Where $\Delta r_{jqk}$ indexes the sampling offset. It takes ten times fewer training epochs than a simple DETR network. The complexity of self-attention becomes $O (w_ih_ic_i^2)$, which is linear complexity according to spatial size $h_i w_i$. The complexity of the cross-attention in decoder becomes $O (NK c_i^2)$ which is independent of spatial size $h_i w_i$.
 In Figure~\ref{fig:deformable-up-efficient}, the top right block indicates deformable attention module in Deformable-DETR.
 
\noindent\textbf{Multi-Scale Feature Maps:}~High-resolution input image features increase the network efficiency, specifically for small objects. However, this is computationally expensive. Deformable-DETR provides high-resolution features without affecting the computation. It uses a feature pyramid containing high and low-resolution features rather than the original high-resolution input image feature map. This feature pyramid has an input image resolution of 1/8, 1/16, and 1/32 and contains its relative positional embeddings. In short, Deformable-DETR replaces the attention module in DETR with the multi-scale deformable attention module to reduce computational complexity and improves performance.
\subsection{UP-DETR}
Dai et al. \cite{updetr23} proposed a few modifications to pre-train the DETR similar to pre-training transformers in NLP. The random-sized patches from the input image are used as object queries to the decoder as input. The pretraining proposed by UP-DETR helps to detect these random-sized query patches. In Figure~\ref{fig:deformable-up-efficient}, the bottom left block denotes UP-DETR. Two issues are addressed during pretraining: multi-task learning and multi-query localization.

\noindent\textbf{Multi-Task Learning:}~Object detection task combines object localization and classification, while these tasks always have distinct features \cite{robustD4, ReClsLoc3, RevisitSH5}. The patch detection damages the classification features. Multi-task learning by patch feature reconstruction and a frozen pretraining backbone is proposed to protect the classification features of the transformer. The feature reconstruction is given as follows: 
\begin{equation}
 \mathcal{L}_{rec}(f_k, \hat{f}_{\hat{\sigma}(k)} ) = 	\parallel {\frac{f_k}{\parallel f_k \parallel_2}-\frac{\hat{f}_{\hat{\sigma}(k)}}{\parallel \hat{f}_{\hat{\sigma}(k)}\parallel_2}}\parallel_2^2
\end{equation}
Here, the feature reconstruction term is $\mathcal{L}_{rec}$. It is the mean-squared error between $l_2$ (normalized) features of patches obtained from the CNN backbone.

\noindent\textbf{Multi-query Localization:}~The decoder of DETR takes object queries as input to focus on different positions and box sizes. When this object queries number $N$ (typically $N = 100$) is high, a single-query group is unsuitable as it has convergence issues. To solve the multi-query localization problem between object queries and patches, UP-DETR proposes an attention mask and query shuffle mechanism. The number of object queries is divided into $X$ different groups, where each patch is provided to $N/X$ object queries. The Softmax layer of the self-attention module in the decoder is modified by adding an attention mask inspired by \cite{tokenmask4} as follows:
\begin{equation}
P(q_i, k_i )=  Softmax(\frac{q_ik_i^T}{\sqrt{d}} + \textbf{M}) . v_i
\end{equation}
\begin{equation}
\bf M_{k,l} = \begin{cases}
0 & \quad k,l\hspace{.5em}in\hspace{.5em}the\hspace{.5em}same\hspace{.5em} group \\
-\infty & \quad otherwise
\end{cases}
\end{equation}
Where $M_{k,l}$ is the interaction parameter of object query $q_k$ and $q_l$. Though object queries are divided into groups, these queries don't have explicit groups during downstream training tasks. Therefore, these queries are randomly shuffled during pre-training by masking $10\%$ query patches to zero, similar to dropout \cite{Dropout79}. 
\subsection{Efficient-DETR}
The performance of DETR also depends on the object queries as the detection head obtains final predictions from them. However, these object queries are randomly initialized at the start of training. Efficient-DETR \cite{efficientDE} based on DETR and Deformable-DETR examines the randomly initialized object blocks, including reference points and object queries, which is one of the reasons for multiple training iterations. In Figure~\ref{fig:deformable-up-efficient}, the bottom right box shows Efficient-DETR. 

Efficient-DETR has two main modules: a dense module and a sparse module. These modules have the same final detection head. The dense module includes the backbone network, encoder network, and detection head. Following \cite{sparsercnn7}, It generates proposals by a class-specific dense prediction using the sliding window and selects Top-k features as object queries and reference points. Efficient-DETR uses 4-d boxes as reference points rather than 2d centres. The sparse network does the same work as the dense network, except for their output size. The features from the dense module are taken as the initial state of the sparse module, which is considered a good initialization of object queries. Both dense and sparse module use one-to-one assignment rule as in \cite{freeA4, ProAn5, noisyAn8}.
\begin{figure*}[h]
\centering
\includegraphics[width=12cm]{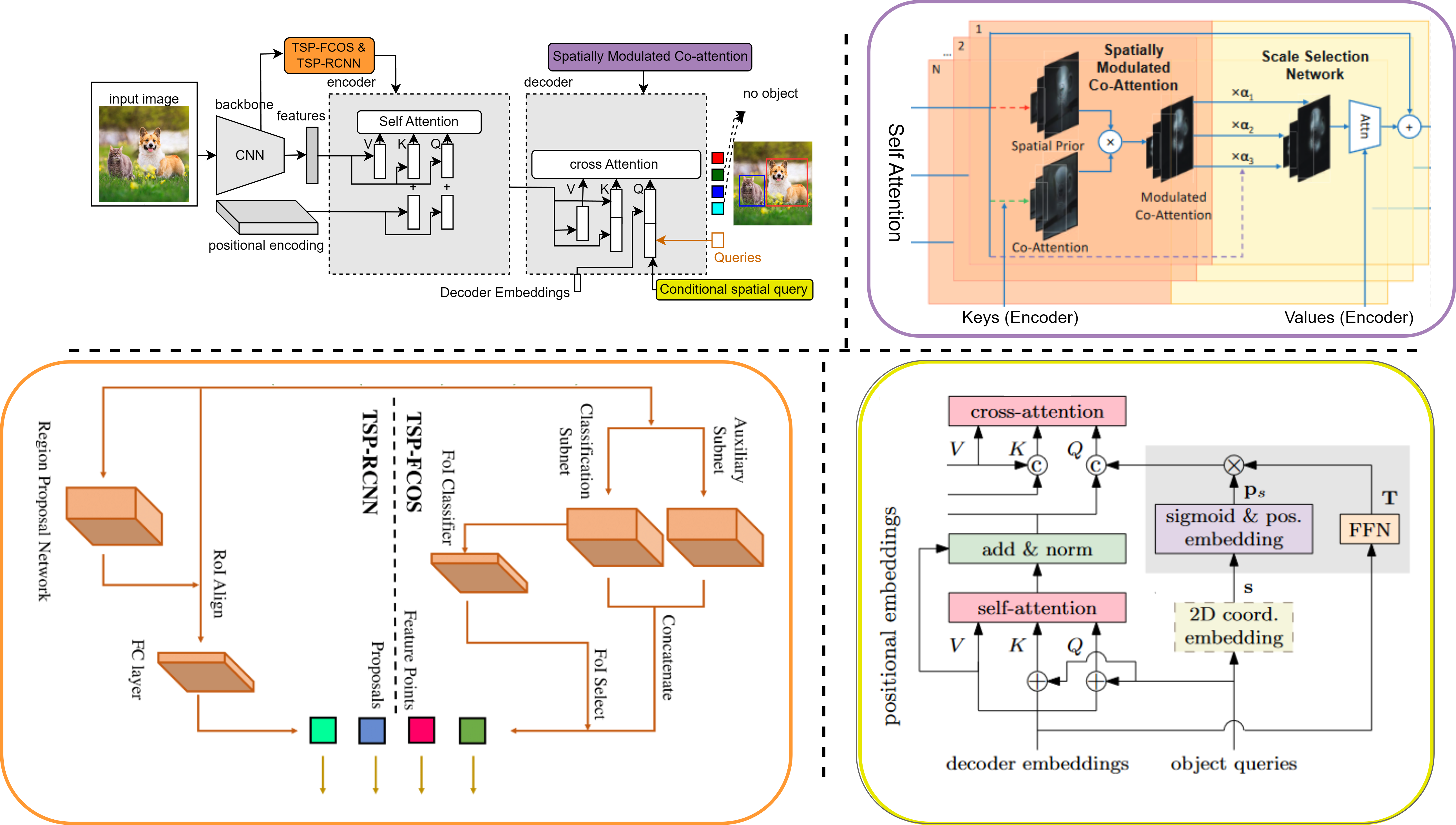}
\caption{The structure of the original DETR after the addition of SMCA-DETR \cite{smca23}, TSP-DETR \cite{tspdetr81} and Conditional-DETR \cite{CondDE}. Here, the top left network is a simple DETR network, along with improvement indicated with small colored boxes. Larger boxes with corresponding colored borders are utilized to illustrate the internal mechanisms of these small colored boxes. The top right block indicates SMCA-DETR, the bottom left block indicates TSP-DETR, and the bottom right box represents Conditional-DETR (images from \cite{smca23, tspdetr81, CondDE}).}\label{fig:SMCA-TSP-conditional}
\end{figure*}
\subsection{SMCA-DETR}
The decoder of the DETR takes object queries as input that are responsible for object detection in various spatial locations. These object queries combine with spatial features from the encoder. The co-attention mechanism in DETR involves computing a set of attention maps between the object queries and the image features to provide class labels and bounding box locations. However, the visual regions in the decoder of DETR related to object query might be irrelevant to the predicted bounding boxes. This is one of the reasons that DETR needs many training epochs to find suitable visual locations to identify corresponding objects correctly. Gao et al. \cite{smca23} introduced a Spatially-Modulated Co-attention (SMCA) module that replaces the existing co-attention mechanism in DETR to overcome the slow training convergence of DETR. In Figure~\ref{fig:SMCA-TSP-conditional}, the top right block represents SMCA-DETR. The object queries estimate the scale and center of its corresponding object, which are further used to set up a 2D  spatial weight map. The initial estimate of scale $l_{h_i},l_{w_i}$  and center $e_{h_i}, e_{w_i}$ of Gaussian-like distribution for object queries $q$ is given as follows:
\begin{align}
e_{h_i}^{nrm}, e_{w_i}^{nrm} &=sigmoid (MLP(q)), \\
l_{h_i},l_{w_i} &= FC (q) \label{eqn:pe2}
\end{align}
Where object query $q$ provides a prediction center in normalized form by sigmoid activation function after two layers of $MLP$. These predicted centers are unnormalized to get the input image's center coordinates $e_{h_i}$ and $e_{w_i}$. The object query also estimates the object scales as $l_{h_i}$, $l_{w_i}$. After the prediction of the object scale and center, SMCA provides a Gaussian-like weight map as follows: 
\begin{equation}
\textbf{W}(x, y )=  exp\left(-\frac{(x-e_{w_i})^2}{\beta l_{w_i}^2}-\frac{(y-e_{h_i})^2}{\beta l_{h_i}^2}\right)
\end{equation}
Where $\beta$ is the hyper-parameter to regulate the bandwidth, $(x, y)$ is the spatial parameter of weight map W. It provides high attention to spatial locations closer to the center and low attention to spatial locations away from the center.
\begin{equation}
A_i=  Softmax(\frac{q_ik_i^T}{\sqrt{d}} + \textbf{log W})v_i 
\end{equation}
Here, $A_i$ is the co-attention map. The difference between the co-attention module in DETR and this co-attention module is the addition of the logarithm of the spatial-map W. The decoder attention network has more attention near predicted box regions, which limits the search locations and thus converges the network faster. 
\subsection{TSP-DETR}
TSP-DETR \cite{Reth78} deals with the cross-attention and the instability of bipartite matching to overcome the slow training convergence of DETR. TSP-DETR proposes two modules based on an encoder network with feature pyramid networks (FPN) \cite{fpn49} to accelerate the training convergence of DETR. In Figure~\ref{fig:SMCA-TSP-conditional}, the bottom left block indicates TSP-DETR. These modules are TSP-FCOS and TSP-RCNN, which used classical one-stage detector FCOS \cite{fcos46} and classical two-stage detector Faster-RCNN \cite{faster23}, respectively. TSP-FCOS used a new Feature of Interest (FoI) module to handle the multi-level features in the transformer encoder. Both modules use the bipartite matching mechanism to accelerate the training convergence. \\

\noindent\textbf{TSP-FCOS:}~The TP-FCOS module follows the FCOS \cite{fcos46} for designing the backbone and FPN \cite{fpn49}. Firstly, the features extracted by the CNN backbone from the input image are fed to the FPN component to produce multi-level features. Two feature extraction heads, the classification head and the auxiliary head, use four convolutional layers and group normalization \cite{GroupNorm7}, which are shared across the feature pyramid stages. Then, the FoI classifier filters the concatenated output of these heads to select top-scored features. Finally, the transformer encoder network takes these FoIs and their positional encodings as input, providing class labels and bounding boxes as output.

\noindent\textbf{TSP-RCNN:}~Like TP-FCOS, this module extracts the features by the CNN backbone and produces multi-level features by the FPN component. In place of two feature extraction heads used in TSP-FCOS, the TSP-RCNN module follows the design of Faster R-CNN \cite{faster23}. It uses Region Proposal Network (RPN) to find Regions of Interest (RoIs) to refine further. Each RoI in this module has an objectness score as well as a predicted bounding box. The RoIAlign \cite{mask-rcnn84} is applied on multi-level feature maps to take RoIs information. After passing through a fully connected network, these extracted features are fed to the Transformer encoder as input. The positional info of these RoI proposals is the four values $(c_{nx}, c_{ny}, w_n, h_n)$, where $(c_{nx}, c_{ny}) \in [0, 1]^2$ represents the normalized value of center and $(w_n, h_n) \in [0, 1]^2$ represents the normalized value of height and width. Finally, the transformer encoder network inputs these RoIs and their positional encoding for accurate predictions.
The FCOS and RCNN modules in TSP-DETR accelerate the training
convergence and improve the performance of the DETR network.
\subsection{Conditional-DETR}
The cross-attention module in the DETR network needs high-quality input embeddings quality to predict accurate bounding boxes and class labels. The high-quality content embeddings increase the training convergence difficulty. Conditional-DETR \cite{CondDE} presents a conditional cross-attention mechanism to solve the training convergence issue of DETR. It differs from the simple DETR by input keys $k_i$ and input queries $q_i$ for cross-attention. In Figure~\ref{fig:SMCA-TSP-conditional}, the bottom right box represents conditional-DETR. The conditional queries are obtained from 2D coordinates along with the embedding output of the previous decoder layer. The predicted candidate box from decoder-embedding is as follows:
\begin{equation}
box=  sig(FFN(e) + [r^T 0 0 ]^T)
\end{equation}
Here, e is the input embedding that is fed as input to the decoder. The $box$ is a 4D vector $[box_{cx} box_{cy} box_{w} box_{h}]$, having the box center value as $(box_{cx},box_{cy})$, width value as $box_{w}$ and height value as $box_{h}$ . $sig()$ function normalizes the predictions varies from 0 to 1. $FFN()$ predicts the un-normalized box. r is the un-normalized 2D coordinate of the reference-point, and $(0,0)$ is the simple DETR. This work either learns the reference point r for each box or generates them from the respective object query. It learns queries for multi-head cross-attention from input embeddings of the decoder. This spatial query makes the cross-attention head consider the explicit region, which helps to localize the different regions for class labels and bounding boxes by narrowing down the spatial range.  
\subsection{WB-DETR}
\begin{figure*}[h]
\centering
\includegraphics[width=16cm]{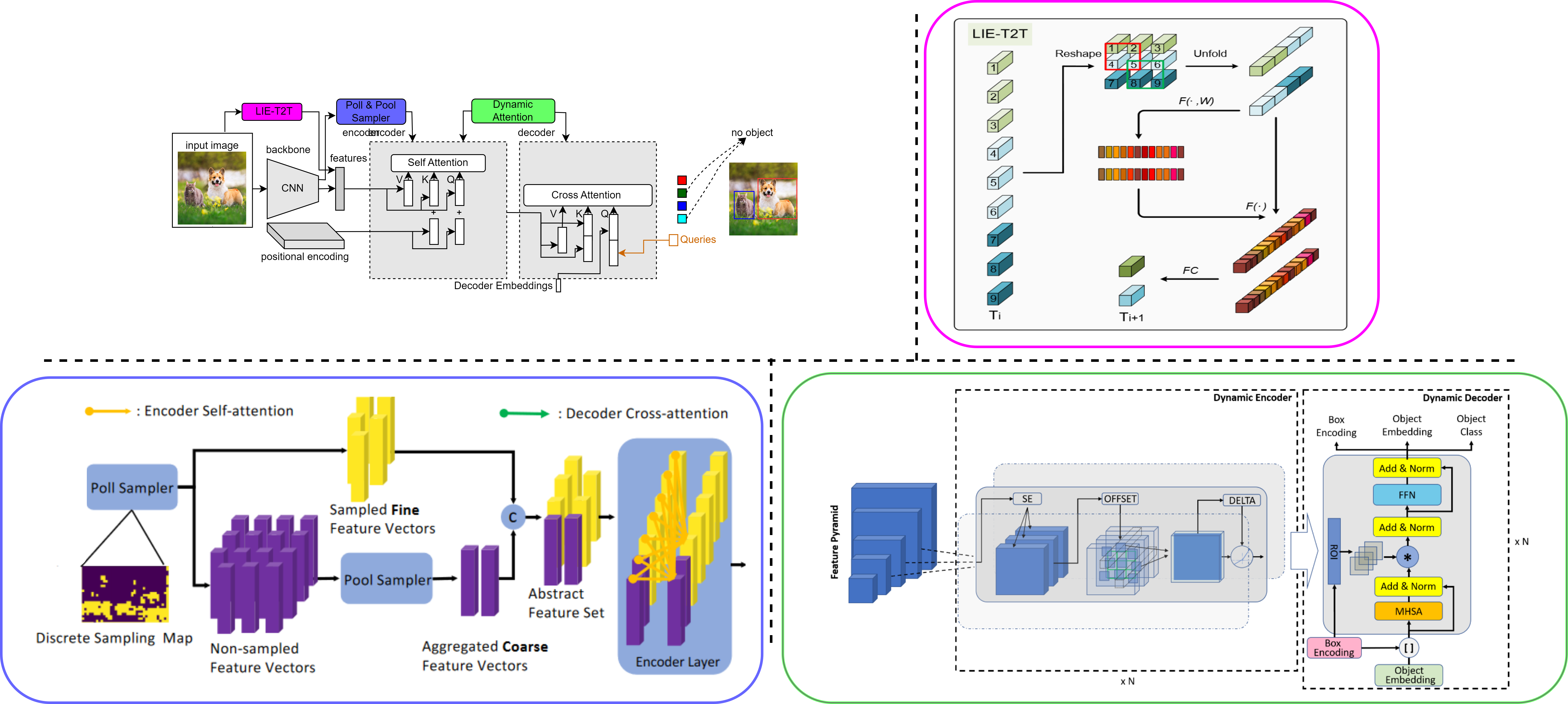}
\caption{ The structure of the original DETR after the addition of WB-DETR \cite{WBdetr4}, PnP-DETR \cite{pnp6} and Dynamic-DETR \cite{DynamicDE}. Here, the top left network is a simple DETR network, along with improvement indicated with small colored boxes. Larger boxes with corresponding colored borders are utilized to illustrate the internal mechanisms of these small colored boxes. The top right block indicates WB-DETR, the bottom left block indicates PnP-DETR, and the bottom right box represents Dynamic-DETR (images from \cite{WBdetr4, pnp6, DynamicDE}).}\label{fig:WB-PnP-dynamic}
\end{figure*}

DETR extracts local features by CNN backbone and gets global contexts by an encoder-decoder network of the transformer. WB-DETR \cite{WBdetr4} proves that the CNN backbone for feature extraction in detection transformers is not compulsory. It contains a transformer network without a backbone. It serializes the input image and feeds the local features directly in each independent token to the encoder as input. The transformer self-attention network provides global information, which can accurately get the contexts between input image tokens. However, the local features of each token and the information between adjacent tokens need to be included as the transformer lacks the ability of local feature modeling. The LIE-T2T (Local Information Enhancement-T2T) module solves this issue by reorganizing and unfolding the adjacent patches and focusing on each patch's channel dimension after unfolding. In Figure~\ref{fig:WB-PnP-dynamic}, the top right block denotes the LIE-T2T module of WB-DETR. The iterative process of the LIE-T2T module is as follows:
\begin{equation}
P = stretch (reshape (Pi))
\end{equation}
\begin{equation}
Q = sig(e_2 \cdot ReLU (e_1 \cdot P )) 
\end{equation}
\begin{equation}
P_{i+1} = e_3 \cdot (P \cdot Q)
\end{equation}
Where $reshape$ function reorganizes $(l_1 \times c_1)$ patches into $(h_i \times w_i \times c_i)$ feature maps. The term $stretch$ denotes unfolding $(h_i \times w_i \times c_i)$ feature maps to $(l_2 \times c_2)$ patches. Here, the fully connected layer parameters are $e_1$, $e_2$, and $e_3$. The $ReLU$ activation is its nonlinear map function, and the $sig$ generates final attention. The channel attention in this module provides local information as the relationship between the channels of the patches is the same as the spatial relation in the pixels of the feature maps.
\subsection{PnP-DETR}
The transformer processes the image feature maps that are transformed into a one-dimensional feature vector to produce the final results. Although effective, using the full feature map is expensive because of useless computation on background regions. PnP-DETR \cite{pnp6} proposes a poll and pool (PnP) sampling module to reduce spatial redundancy and make the transformer network computationally more efficient. This module divides the image feature map into contextual background features and fine foreground object features. Then, the transformer network uses these updated feature maps and translates them into the final detection results. In Figure~\ref{fig:WB-PnP-dynamic}, the bottom left block indicates PnP-DETR. This PnP Sampling module includes two types of samplers: a pool sampler and a poll sampler, as explained below.\\
\noindent\textbf{Poll Sampler:}~The poll sampler provides fine feature vectors $\mathds{V}_f$. A meta-scoring module is used to find the informational value for every spatial location (x, y):
\begin{equation}
a_{xy} = ScoreNet(v_{xy} , \theta s)
\end{equation}
The score value is directly related to the information of feature vector $v_{xy}$. These score values are sorted as follows: 
\begin{equation}
[a_z, |z = 1, . . . , Z], \aleph= Sort({a_{xy} })
\end{equation}
Where $Z = h_iw_i$ and $\aleph$ is the sorting order. The top $N_s$-scoring vectors are selected to get fine features:
\begin{equation}
\mathds{V}_f = [v_z , |z = 1, . . . , N_s ] 
\end{equation}
Here, the predicted informative value is considered as a modulating factor to sample the fine feature vectors:
\begin{equation}
\mathds{V}_f = [v_z \times a_z , |z = 1, . . . , N_s ] 
\end{equation}
To make the learning stable, the feature vectors are normalized:
\begin{equation}
\mathds{V}_f = [L_{norm}(v_z) \times a_z, |z = 1, . . . , N_s ] 
\end{equation}
Here, $L_{norm}$ is the layer normalization, $N_s = \alpha Z$, where $\alpha$ is the poll ratio factor. This sampling module reduces the training computation.\\
\noindent\textbf{Pool Sampler:}~The poll sampler gets the fine features of foreground objects. A pool sampler compresses the background region's remaining feature vectors that provide contextual information. It performs weighted pooling to get a small number of background features $M_b$ motivated by double attention operation \cite{dan9} and bilinear pooling \cite{bilinear27}. The remaining feature vectors of the background region are:
\begin{equation}
\mathds{V}_b = \mathds{V} \backslash \mathds{V}_f = \{\textbf{v}_b ,|b = 1, . . . , Z-N \}
\end{equation}
The aggregated weights $\bf{a}_b\in\mathds{R}^{M_b} $are obtained by projecting the features with weight values $\textbf{w}^s \in \mathds{R}^{c_i \times M_b} $ as:
\begin{equation}
\textbf{a}_b = \textbf{v}_b \textbf{w}^s
\end{equation}
The projected features with learnable weight $\textbf{w}^p \in \mathds{R}^{c_i \times c_i}$ are obtained as follows:
\begin{equation}
\Acute{\textbf{v}}_b = \bf{v}_b \bf{w}^p
\end{equation}
The aggregated weights are normalized over the non-sampled regions with Softmax as follows:
\begin{equation}
a_{bm} = \frac{e^a_{bm}}{\sum_{\Acute{b}=1}^{N-Z}e^a~\Acute{b}~m }
\end{equation}
By using the normalized aggregation weight, the new feature vector is obtained that provides information of non-sampled regions: 
\begin{equation}
\textbf{v}_m = \sum_{b=1}^{Z-N} \Acute{\textbf{v}}_b \times a_{bm}
\end{equation}
By considering all Z aggregation weights, the coarse background contextual feature vector is as follows:
\begin{equation}
\mathds{V}_c = \{\textbf{v}_m, |b = 1, . . . , M_b \} 
\end{equation}
The pool sampler provides context information at different scales using aggregation weights. Here, some feature vectors may provide local context while others may capture global context. 
\subsection{Dynamic-DETR}
Dynamic-DETR \cite{DynamicDE} introduces dynamic attention in the encoder-decoder network of DETR to solve the slow training convergence issue and detection of small objects. Firstly, a convolutional dynamic encoder is proposed to have different attention types to the self-attention module of the encoder network to make the training convergence faster. The attention of this encoder depends on various factors such as spatial effect, scale effect and input feature dimensions effect. Secondly, ROI-based dynamic attention is replaced with cross-attention in the decoder network. This decoder helps to focus on small objects, reduces learning difficulty and converges the network faster. In Figure~\ref{fig:WB-PnP-dynamic}, the bottom right box represents Dynamic-DETR. This dynamic encoder-decoder network is explained in detail as follows.\\
\noindent\textbf{Dynamic Encoder:}~The Dynamic-DETR uses a convolutional approach for the self-attention module. Given the feature vectors $F = \{F1, \cdot\cdot \cdot , F_n\}$, where n=5 represents object detectors from the feature pyramid, the multi-scale self-attention (MSA) is as follows: 
\begin{equation}
Attn = MSA (F).F
\end{equation}
However, it is impossible because of the various scale feature map from the FPN. The feature maps of different scales are equalized within neighbours using 2D convolution as in the Pyramid Convolution \cite{se76}. It focuses on spatial locations of the un-resized mid-layer and transfers information to its scaled neighbours. Moreover, SE\cite{se07} is applied to combine the features to provide scale attention. 

\noindent\textbf{Dynamic Decoder:}~The dynamic decoder uses mixed attention blocks in place of multi-head layers to ease the learning in the cross-attention network and improves the detection of small objects. It also uses dynamic convolution instead of a cross-attention layer inspired by ConvBERT \cite{bert96} in natural language processing (NLP). Firstly, RoI Pooling \cite{faster23} is introduced in the decoder network. Then position embeddings are replaced with box encoding $BE \in \mathds{R}^{p \times 4}$ as the image size. The output from the dynamic encoder, along with box encoding $BE$, is fed to the dynamic decoder to pool image features  $R \in \mathds{R}^{p \times s \times s \times c_i}$ from feature pyramid as follows:
\begin{equation}
R = RoI_{pool}(F_{encoder}, BE, s) 
\end{equation}
where s is the size of pooling parameter, $c_i$ represents quantity of channels of $F_{encoder}$. To feed this in the cross-attention module,
input embeddings $qe \in R^{p \times c_i}$ are required for object queries.
These embeddings are passed through the Multi-Head self Attention (MHSAttn) layer as:
\begin{equation}
qe^\ast = MHSAttn(qe, qe, qe)
\end{equation}
Then these query embeddings are passed through fully-
connected layer (dynamic filters) as follows:
\begin{equation}
Filter^{qe} = FC(qe^\ast)
\end{equation}
Finally, cross-attention between features and object queries is performed with 1 × 1 convolution using dynamic filters $Filter^{qe}$:
\begin{equation}
 qe^F= Con_{1\times 1}(F, Filter^{qe})
\end{equation}
These features are passed through the FFN layers to provide various predictions as updated object-embedding, updated box-encoding, and the object class. This process eases the learning of the cross-attention module by focusing on sparse areas and then spreading to global regions.
\begin{figure*}[h]
\centering
\includegraphics[width=16cm]{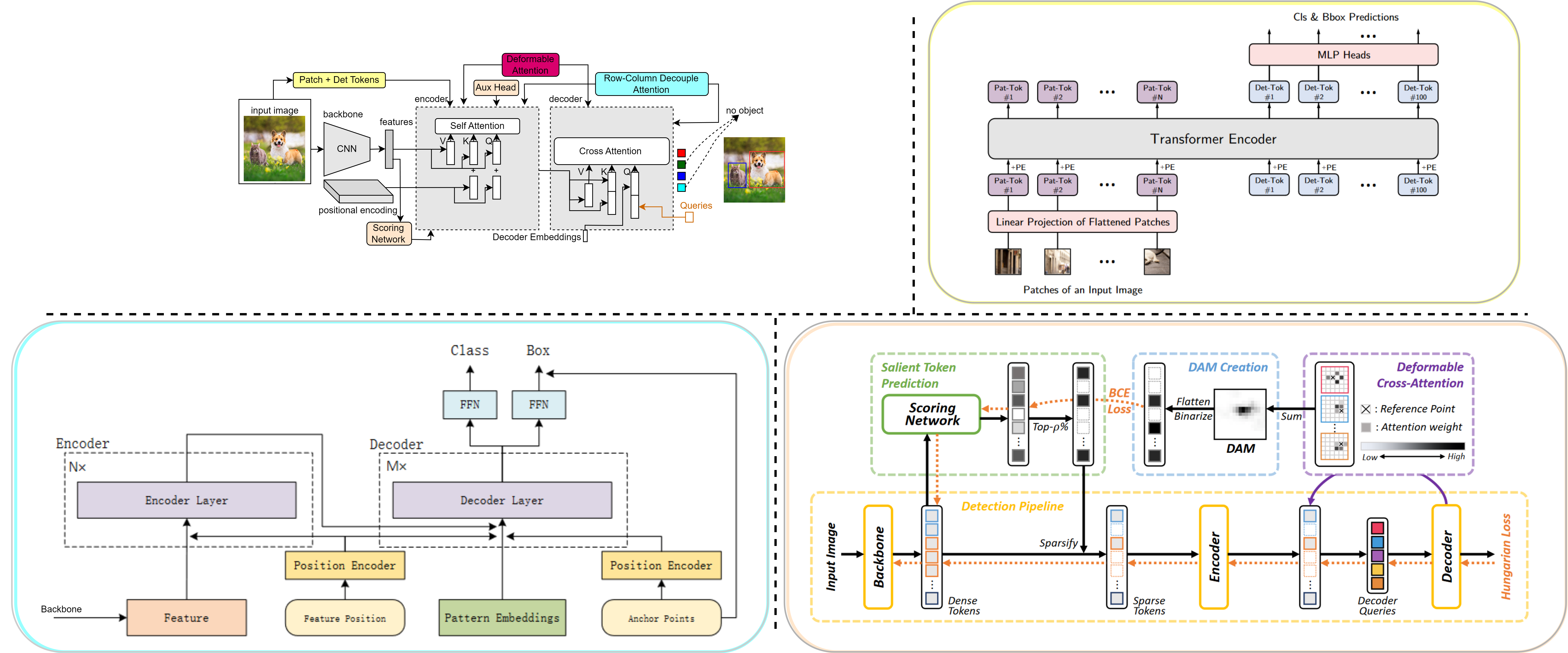}
\caption{The structure of the original DETR after the addition of YOLOS-DETR \cite{yolos6}, Anchor-DETR \cite{Anchor-detr} and Sparse-DETR \cite{sparsedetr}. Here, the top left network is a simple DETR network, along with improvement indicated with small colored boxes. Larger boxes with corresponding colored borders are utilized to illustrate the internal mechanisms of these small colored boxes. The top right block indicates YOLOS-DETR, the bottom left block indicates Anchor-DETR, and the bottom right box represents Sparse-DETR (images from \cite{yolos6, Anchor-detr, sparsedetr} ).}\label{fig:yolos-anchor-sparse}
\end{figure*}
\subsection{YOLOS-DETR}
Vision Transformer (ViT) \cite{ViT23} inherited from NLP performs well on the image recognition task. ViT-FRCNN \cite{ViTFrcnn2} uses a pre-trained backbone (ViT) for a CNN-based detector. It utilizes convolution neural networks and relies on strong 2D inductive biases and region-wise pooling operations for object-level perception. Other similar works, such as DETR \cite{detr34}, introduce 2D inductive bias using CNNs and pyramidal features. YOLOS-DETR \cite{yolos6} presents the transferability and versatility of the Transformer from image recognition to detection in the sequence aspect using the least information about the spatial design of the input. It closely follows ViT architecture with two simple modifications. Firstly, it removes the image-classification patches [CLS] and adds randomly initialized one hundred detection patches [DET] as \cite{AutoAssign6} along with the input patch embeddings for object detection. Secondly, similar to DETR, a bipartite matching loss is used instead of ViT classification loss. The transformer encoder takes the generated sequence as input as follows: 
\begin{equation}
\begin{split}
s_0 = [\textbf{I}^1_{p}\textbf{L}; \cdot \cdot \cdot; \textbf{I}^M_{p}\textbf{L}; \textbf{I}^1_{d}; \cdot\cdot\cdot ; \textbf{I}^{100}_{d}] + \textbf{PE}
\end{split}
\end{equation}
Where, I is the input image $\textbf{I} \in \mathds{R}^{h_i \times w_i \times c_i} $ that is reshaped into 2D tokens $\textbf{I}_{p} \in \mathds{R}^{n_i \times (R^2 \cdot c_i)}$. Here, $h_i$ represents the height, and $w_i$ indicates the width of the input image. $c_i$ is the total channels. $(r,r)$ is each token resolution, $n_i = \frac{h_iw_i}{r^2}$ is the total number of tokens. These tokens are mapped to $D_i$ dimensions with linear projection $\textbf{L} \in \mathds{R}^{(r^2 \cdot c_i)\times D_i} $. The result of this projection is $\textbf{I}_{p}\textbf{L}$. The encoder also takes randomly initialized one hundred learnable tokens $\textbf{I}_{d} \in \mathds{R}^{100 \times D_i} $. To keep the positional information, positional embeddings $\textbf{PE} \in \mathds{R}^{(n_i+100) \times D_i}$ are also added. The encoder of the transformer contains a multi-head self-attention mechanism and one MLP block with GELU \cite{geluu6} non-linear activation function. The Layer Normalization (LN) \cite{ln78} is added between each self-attention and MLP block as follows:
\begin{align}
\Acute{s}_n &\quad = MHSAttn(LN (s_{n-1})) + s_{n-1}\\
s_n &\quad = MLP (LN (\Acute{s}_{n})) + \Acute{s}_{n}
\end{align}
Where $s_n$ is the encoder input sequence. In Figure~\ref{fig:yolos-anchor-sparse}, the top right block indicates YOLOS-DETR.
\subsection{Anchor-DETR}
DETR uses learnable embeddings as object queries in the decoder network. These input embeddings do not have a clear physical meaning and cannot illustrate where to focus. It is challenging to optimize the network as object queries concentrate on something other than specific regions. Anchor-DETR \cite{Anchor-detr} solves this issue by proposing object queries as anchor points that are extensively used in CNN-based object detectors. This query design can provide multiple object predictions at one region. Moreover, a few modifications in the attention are proposed that reduce the memory cost and improve performance. In Figure~\ref{fig:yolos-anchor-sparse}, the bottom left block shows Anchor-DETR. The two main contributions of Anchor-DETR: query and attention variant design, are explained as follows:

\noindent\textbf{Row and Column Decoupled-Attention:}~DETR requires huge GPU memory as in \cite{luna87, attnlinear3} because of the complexity of the cross-attention module. It is more complex than the self-attention module in the decoder. Although Deformable-DETR reduces memory cost, it still causes random memory access, making the network slower. Row-Column Decoupled Attention (RCDA), as shown in the bottom left block of Figure~\ref{fig:yolos-anchor-sparse}, reduces memory and provides similar or better efficiency.

\noindent\textbf{Anchor Points as Object Queries:}~The CNN-based object detectors consider anchor points as the relative position of the input feature maps. In contrast, transformer-based detectors take uniform grid locations, hand-craft locations, or learned locations as anchor points. Anchor-DETR considers two types of anchor points: learned anchor locations and grid anchor locations. The gird anchor locations are input image grid points. The learned anchor locations are uniform distributions from 0 to 1 (randomly initialized) and updated using the learned parameters. 
\subsection{Sparse-DETR}
\begin{figure*}[h]
\centering
\includegraphics[width=15cm]{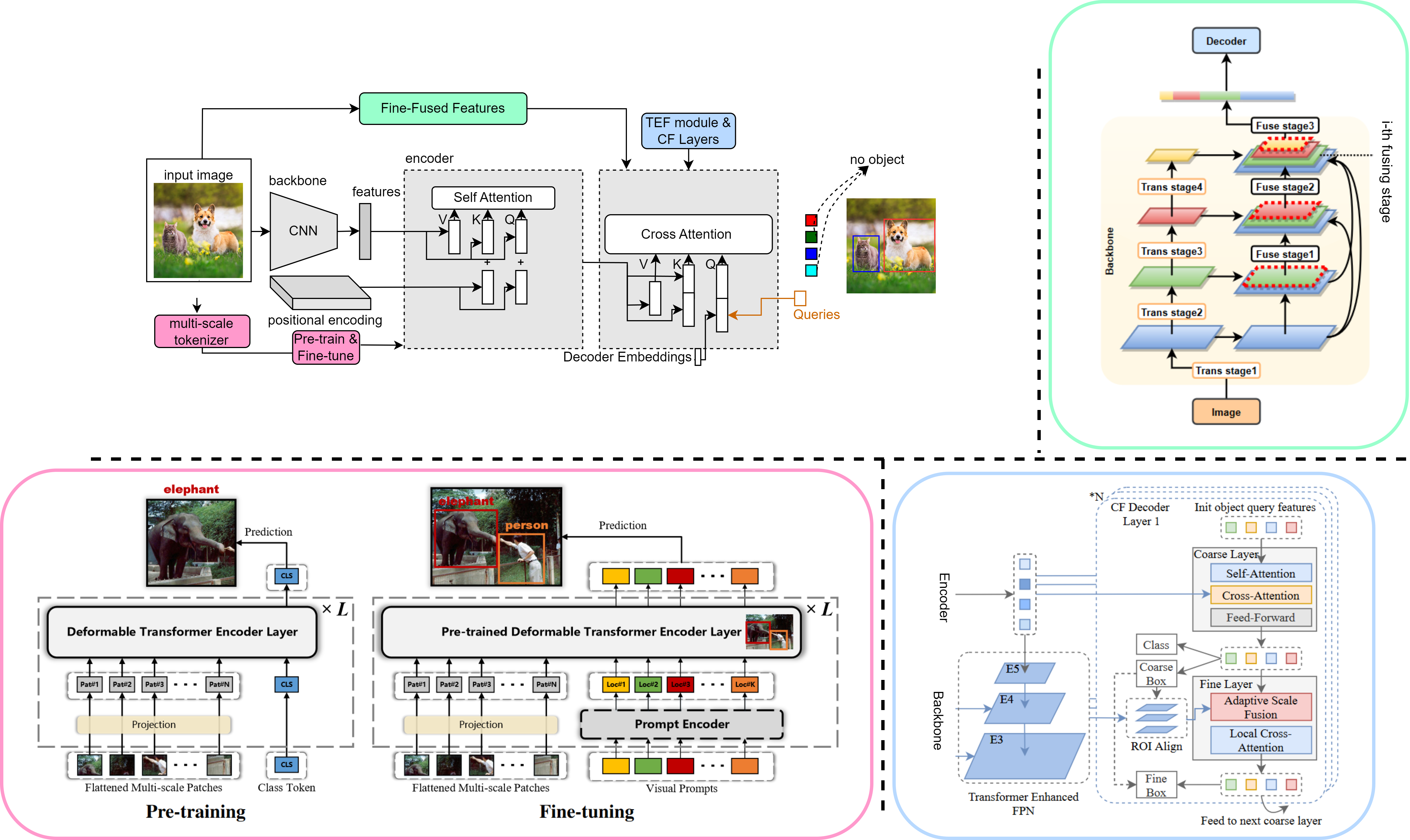}
\caption{The structure of the original DETR after the addition of D$^2$ETR \cite{decoderonly60}, FP-DETR \cite{fpdetr} and CF-DETR \cite{cfdetr}. Here, the top left network is a simple DETR network, along with improvement indicated with small colored boxes. Larger boxes with corresponding colored borders are utilized to illustrate the internal mechanisms of these small colored boxes. The top right block indicates D$^2$ETR, the bottom left block indicates FP-DETR, and the bottom right box represents CF-DETR (images from \cite{decoderonly60, fpdetr, cfdetr}).}\label{fig:D-only-FP-CF}
\end{figure*}
Sparse-DETR \cite{sparsedetr} filters the encoder tokens by learnable cross-attention map predictor. After distinguishing these tokens in the decoder network, it focuses only on foreground tokens to reduce computational costs.

Sparse-DETR introduces the scoring module, aux-heads in the encoder, and the Top-k queries selection module for the decoder. In Figure~\ref{fig:yolos-anchor-sparse}, the bottom right box represents Sparse-DETR. Firstly, it determines the saliency of tokens, fed as input to the encoder, using the scoring network that selects top $\rho\%$ tokens. Secondly, the aux-head takes the top-k tokens from the output of the encoder network. Finally, the top-k tokens are used as the decoder object queries. The salient token prediction module refines encoder tokens that are taken from the backbone feature map using threshold $\rho$ and updates the features $x_l-1$ as:
\begin{equation}
\bf x_{l}^m = \begin{cases}
x_{l-1}^m &  \quad m \notin \Omega_r^q  \\
LN(FFN(y_l^m)+y_l^m) & \quad m \in \Omega_r^q, \nonumber
\end{cases}
\end{equation}
\begin{equation}
where \quad y_l^m = LN (DeformAttn(x_{l-1}^m, x_{l-1}) + x_{l-1}^m)
\end{equation}
Where DeformAttn is the deformable attention, FFN is the Feed-Forward Network, and LN is the Layer-Normalization. Then, the Decoder Cross-Attention Map (DAM) accumulates the attention weights of decoder object queries, and the network is trained  by minimizing loss between prediction and binarized DAM as follows:
\begin{equation}
\mathcal{L}_{dam} = \frac{-1}{M}\sum_{k=1}^{M} BCELoss(sn(x_{f}),DAM_k^b)
\end{equation}
Where BCELoss is the binary cross-entropy (BCE) loss,  $DAM_k^b$ is the k-th binarized DAM value of the encoder token, and $sn$ is the scoring network. In this way, sparse-DETR minimizes the computation by significantly eliminating encoder tokens. 
\subsection{D$^2$ETR}
Much work \cite{ Reth78, Deformable54, efficientDE, CondDE, smca23} has been proposed to make the training convergence faster by modifying the cross-attention module. Many researchers \cite{Deformable54} used multi-scale feature maps to improve performance for small objects. However, the solution for high computation complexity has yet to be proposed. D$^2$ETR \cite{decoderonly60} achieves better performance with low computational cost. Without an encoder module, the decoder directly uses the fine-fused feature maps provided by the backbone network with a novel cross-scale attention module. The D$^2$ETR contains two main modules a backbone and a decoder. The backbone network based on  Pyramid Vision Transformer (PVT) consists of two parallel layers, one for cross-scale interaction and another for intra-scale interaction. This backbone contains four transformer levels to provide multi-scale feature maps. All levels have the same architecture depending on the basic module of the selected Transformer. The backbone also contains three fusing levels in parallel with four transformer levels. These fusing levels provide a cross-scale fusion of input features. The i-th fusing level is shown in the top right block of Figure~\ref{fig:D-only-FP-CF}. The cross-scale attention is formulated as follows:
\begin{equation}
f_j = \textbf{L}_j (f_{j-1})
\end{equation}
\begin{equation}
f_j^\ast = SA (f_q, f_k, f_v)
\end{equation}
\begin{equation}
f_q = f_j, f_k= f_v =  [f_1^\ast, f_2^\ast, ... , f_{j-1}^\ast, f_j]
\end{equation}
where $f_j^\ast$ the fused form feature map $f_j$. Given that L is the input of the decoder as the last-level feature map, the final result of cross-scale attention is $f_1^\ast, f_2^\ast, ..., f_{L}^\ast$. The output of this backbone is fed to the decoder that takes object queries in parallel. It provides output embeddings independently transformed into class labels and box coordinates by a forward feed network. Without an encoder module, the decoder directly used the fine-fused feature maps provided by the backbone network with a novel cross-scale attention module providing better performance with low computational cost.
\subsection{FP-DETR}
Modern CNN-based detectors like YOLO \cite{yolox34} and Faster-RCNN \cite{faster23} utilize specialized layers on top of backbones pre-trained on ImageNet to enjoy pre-training benefits such as improved performance and faster training convergence. DETR network and its improved version \cite{updetr23} only pre-train its backbone while training both encoder and decoder layers from scratch. Thus, the transformer needs massive training data for fine-tuning. The main reason for not pre-training the detection transformer is the difference between the pre-training and final detection tasks. Firstly, the decoder module of the transformer takes multiple object queries as input for detecting objects, while ImageNet classification takes only a single query (class token). Secondly, the self-attention module and the projections on input query embeddings in the cross-attention module easily overfit a single class query, making the decoder network difficult to pre-train. Moreover, the downstream detection task focuses on classification and localization, while the upstream task considers only classification for the objects of interest. 

FP-DETR \cite{fpdetr} reformulates the pre-training and fine-tuning stages for detection transformers. In Figure~\ref{fig:D-only-FP-CF}, the bottom left block indicates FP-DETR. It takes only the encoder network of the detection transformer for pre-training as it is challenging to pre-train the decoder on the ImageNet classification task. Moreover, DETR uses both the encoder and  CNN backbone as feature extractors. FP-DETR replaces the CNN backbone with a multi-scale tokenizer and uses the encoder network to extract features. It fully pre-trains the Deformable-DETR on the ImageNet dataset and fine-tunes it for final detection that achieves competitive performance. 
\subsection{CF-DETR}
CF-DETR \cite{cfdetr} observes that COCO-style metric Average Precision (AP) results for small objects on detection transformers at low IoU threshold values are better than CNN-based detectors. It refines the predicted locations by utilizing local information, as incorrect bounding box location reduces performance on small objects. CF-DETR introduces the Transformer Enhanced FPN (TEF) module, coarse layers and fine layers in the decoder network of DETR. In Figure~\ref{fig:D-only-FP-CF}, the bottom right box represents CF-DETR. The TEF module provides the same functionality as FPN, have non-local features E4 and E4 extracted from the backbone, and  E5 features taken from the encoder output. The features of the TEF module and the encoder network are fed to the decoder as input. The decoder modules introduce a coarse block and a fine block. The coarse block selects foreground features from the global context. The fine block has two modules, Adaptive Scale-Fusion (ASF) and Local Cross-Attention (LCA), further refining coarse boxes. In short, these modules refine and enrich the features by fusing global and local and global information to improve detection transformer performance.
\begin{figure*}[h]
\centering
\includegraphics[width=15cm]{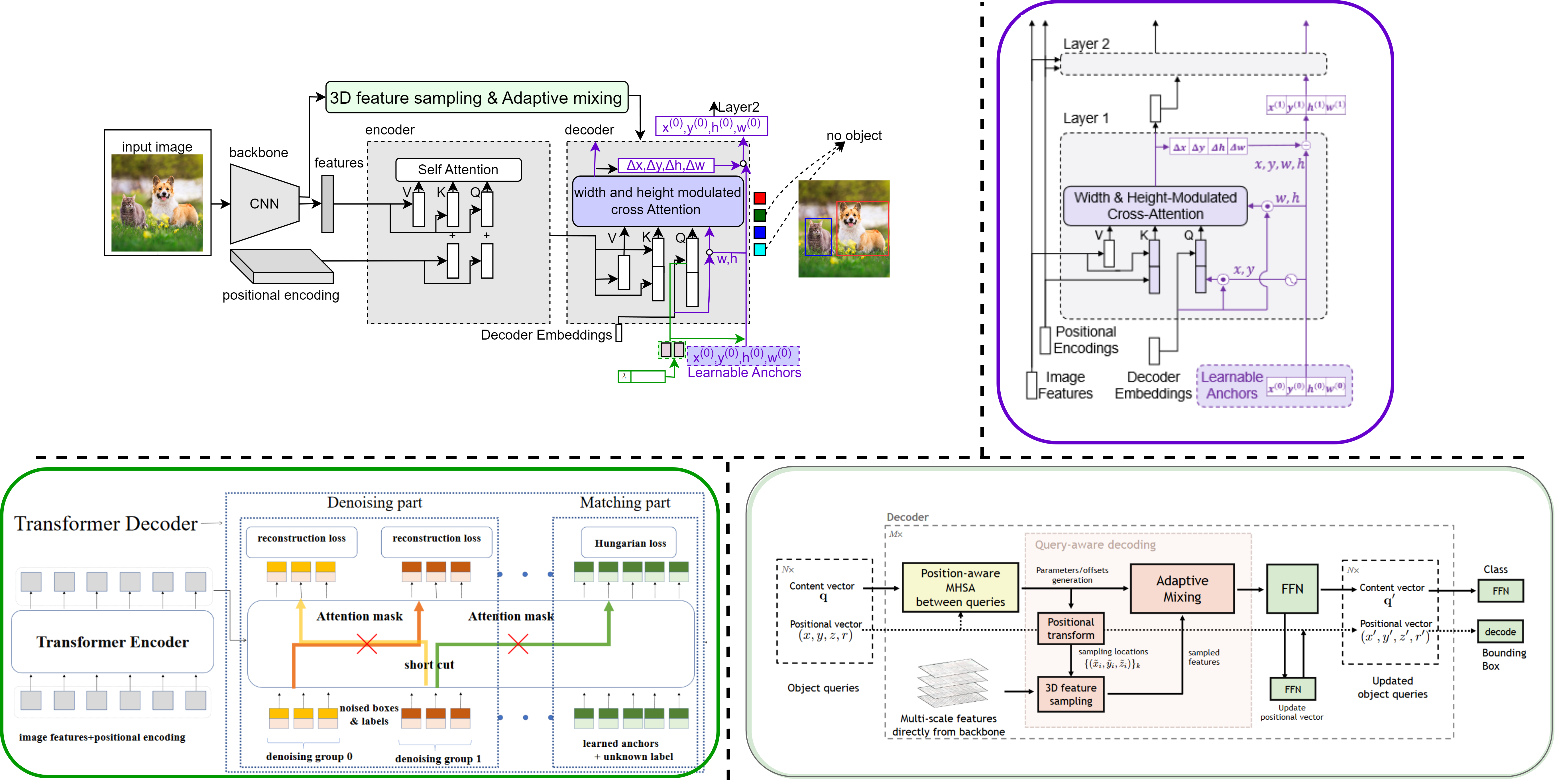}
\caption{The structure of the original DETR after the addition of DAB-DETR \cite{dab89}, DN-DETR \cite{dn42} and AdaMixer \cite{adamixer7}. Here, the top left network is a simple DETR network, along with improvement indicated with small colored boxes. Larger boxes with corresponding colored borders are utilized to illustrate the internal mechanisms of these small colored boxes. The top right block indicates DAB-DETR, the bottom left block indicates DN-DETR, and the bottom right box represents AdaMixer (image from \cite{dab89, dn42, adamixer7}).}\label{fig:DAB-DN-Adamixer}
\end{figure*}
\subsection{DAB-DETR}
DAB-DETR \cite{dab89} uses the bounding box coordinates as object queries in the decoder and gradually updates them in every layer. In Figure~\ref{fig:DAB-DN-Adamixer}, the top right block indicates DAB-DETR. These box coordinates make training convergence faster by providing positional information and using the height and width values to update the positional attention map. This type of object query provides better spatial prior for the attention mechanism and provides a simple query formulation mechanism. 

The decoder network contains two main networks a self-attention network to update queries and a cross-attention network to find features probing. The difference between the self-attention of original DETR and DAB-DETR is that query and key matrices have also position information taken from bounding-box coordinates. The cross-attention module concatenates the position and content information in key and query matrices and determines their corresponding heads. The decoder takes input embeddings as content queries and anchor boxes as positional queries to find object probabilities related to anchors and content queries. This way, dynamic box coordinates used as object queries provide better prediction, making the training convergence faster and increasing detection results for small objects.
\subsection{DN-DETR}
DN-DETR \cite{dn42} uses noised object queries as an additional decoder input to reduce the instability of the bipartite-matching mechanism in DETR, which causes the slow convergence problem. In Figure~\ref{fig:DAB-DN-Adamixer}, the bottom left block indicates DN-DETR. The decoder queries have two parts: the denoising part containing noised ground-truth box-label pairs as input and the matching part containing learnable anchors as input. The matching part $M = \{M_0, M_1, ..., M_{l-1}\}$ determines the resemblance between the ground-truth label pairs and the decoder output, while the denoising part $d = \{ d_0, d_1, ..., d_{k-1}\}$ attempts to reconstruct the ground-truth objects as:
\begin{equation}
 Output = Decoder(d,M,I|A) 
\end{equation}
Where $I$ is the image features taken as input from the transformer encoder, and $A$ is the attention mask that stops the information transfer between the matching and denoising parts and among different noised levels of the same ground-truth objects. The decoder has noised levels of ground-truth objects where noise is added to bounding boxes and class labels, such as label flipping. It contains a hyper-parameter $\lambda$ for controlling the noise level. The training architecture of DN-DETR is based on DAB-DETR, as it also takes bounding box coordinates as object queries. The only difference between these two architectures is the class label indicator as an additional input in the decoder to assist label denoising. The bounding boxes are updated inconsistently in DAB-DETR, making relative offset learning challenging. The denoising training mechanism in DN-DETR improves performance and training convergence. 
\begin{figure*}[h]
\centering
\includegraphics[width=15cm]{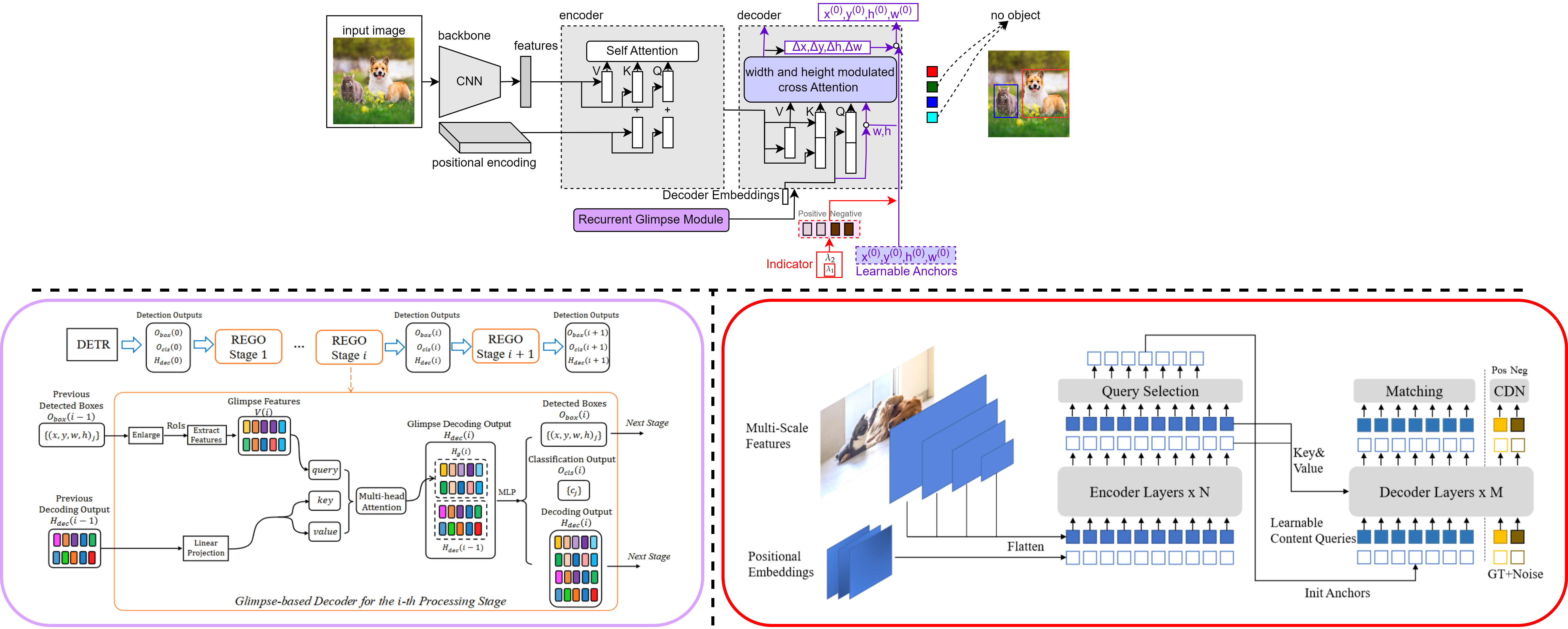}
\caption{The structure of the original DETR after the addition of REGO-DETR \cite{rego2} and DINO \cite{dino23}. Here, the top network is a simple DETR network, along with improvement indicated with small colored boxes. Larger boxes with corresponding colored borders are utilized to illustrate the internal mechanisms of these small colored boxes. The bottom left block indicates REGO-DETR and the bottom right box represents DINO (images from \cite{rego2, dino23}).}\label{fig:Rego-Dino}
\end{figure*}
\subsection{AdaMixer}
AdaMixer \cite{adamixer7} considers the encoder an extra network between the backbone and decoder that limits the performance and slower the training convergence because of its design complexity. AdaMixer provides a detection transformer network without an encoder. In Figure~\ref{fig:DAB-DN-Adamixer}, the bottom right box represents AdaMixer. The main modules of AdaMixer are explained as follows.

\noindent\textbf{3D feature space:}~For 3D feature space, the input feature map from the CNN backbone with the downsampling stride $s_i^{f}$, is first transformed by a linear-layer to the same $d_f$ channel and computed the coordinate of its z-axis as follows:
\begin{equation}
 z^f_i = log_2(s^f_i / s_{b}).
\end{equation}
Where, height $h_i$ and width $w_i$ of feature maps (different strides) is rescaled to $h_i/s_b$ and $w_i/s_b$, where $s_b = 4$.  

\noindent\textbf{3D feature sampling process:}~In the sampling process, the query generates $I_p$ groups of vectors to $I_{p}$ points, ${(\Delta x_j, \Delta y_j, \Delta z_j)}I_{p}$, where each vector is dependent on its content-vector $q_i$ by a linear-layer $L_i$ as follows:
\begin{equation}
{( \Delta x_j, \Delta y_j, \Delta z_j)}I_p = L_i(q_i).
\end{equation}
These offset values are converted into sampling positions w.r.t position vector of object query as follows:
\begin{equation}
\begin{cases}
\tilde{x_j} &= x + \Delta x_j . 2^{z-r} \\
\tilde{y_j} &= y + \Delta y_j . 2^{z+r} , \\
\tilde{z_j} &= z + \Delta z_j,
\end{cases}
\end{equation}
The interpolation over the 3D feature space first samples by bilinear interpolation in the $(x_i, y_i)$ space and then interpolates on the z-axis by Gaussian weighting with weight for the i-th feature map is as follows:
\begin{equation}
\tilde{w}_i = \frac{exp(-(\tilde{z}-z_i^f)^2 / \Gamma_z)}  {\sum_i exp(-(\tilde{z}-z_i^f)^2 / \Gamma_z)}
\end{equation}
where $\Gamma_z $ is the softening coefficient to interpolate values
over the z-axis ($\Gamma_z = 2 $ ). This process makes decoder detection learning easier by taking feature samples according to the query. 

\noindent\textbf{AdaMixer Decoder:}~The decoder module in AdaMixer takes a content vector $q_i$ and positional vector $(x_i, y_i, z_i, r_i)$ as input object queries.
The position-aware multi-head self-attention is applied between these queries as follows.
\begin{equation}
Attn(q_i, k_i, v_i )= Softmax(\frac{q_ik_i^T}{\sqrt{d}} + \alpha X).v_i
\end{equation}
Where $X_{kl}=log(\lvert box_k\cap box_l \slash\lvert box_k\rvert+\epsilon),\epsilon=10^{-7}$. The $X_{kl} = 0 $ indicates the $box_k$ is inside the $box_l$ and $X_{kl} = l$ represents no overlapping between $box_k$ and $box_l$. This position vector is updated at every stage of the decoder network. The AdaMixer decoder module takes a content vector and a positional vector as input object queries. For this, multi-scale features taken from the CNN backbone are converted into 3D feature space as the decoder should consider $(x_i, y_i)$ space as well as adjustable in terms of scales of detected objects. It takes the sampled features from this feature space as input. It applies the adaMixer mechanism to provide final predictions of input queries without using an encoder network to reduce the computational complexity of detection transformers.
\subsection{REGO-DETR}
REGO-DETR \cite{rego2} proposes an RoI-based method for detection refinement to improve the attention mechanism in DETR. In Figure~\ref{fig:Rego-Dino}, the bottom left block denotes REGO-DETR. It contains two main modules: a multi-level recurrent mechanism and a glimpse-based decoder. In the multi-level recurrent mechanism, bounding boxes detected in the previous level are considered to get glimpse features. These are converted into refined attention using earlier attention in describing objects. The k-th processing level is as follows:
\begin{equation}
\begin{cases}
O_{class} (k)= DF_{class}(H_{de}(k)) \\
O_{bbox} (k)= DF_{bbox}(H_{de}(k)) + O_{bbox}(k-1)
\end{cases}
\end{equation}
Where $O_{class}\in\mathds{R}^{ M_d\times M_c}$ and $O_{bbox}\in\mathds{R}^{ M_d\times4}$. Here, $M_d$ and $M_c$ represent the total number of predicted objects and classes, respectively. $DF_{class}$ and $DF_{bbox}$ are functions that convert the input features into desired outputs. $H_{de}(k)$ is the attention of this level after decoding as:
\begin{equation}
H_{de}(k)= [H_{gm}(k), H_{de}(k-1)] \\
\end{equation}
Where $H_{gm}(k)$ is the glimpse features according to $H_{de}(k-1)$ and previous levels. These glimpse features are transformed using multi-head cross-attention into refined attention outputs according to previous attention outputs as:
\begin{equation}
H_{gm}(k)= Attn (V(k), H_{de}(k-1)), \\
\end{equation} 
For extracting the glimpse features $V(k)$, the following operation is performed:
\begin{equation}
V(k) = FE_{ext}( X, RI(O_{bbox}(k-1), \alpha (k))), \\
\end{equation} 
Where $FE_{ext}$ is the feature extraction function, $\alpha (k)$ is a scalar parameter, and RI is the RoI computation. In this way, The Region-of-Interest (RoI) based refinement modules make the training convergence of the detection transformer faster and provide better performance.
\subsection{DINO}
DN-DETR adds positive noise to the anchors taken as object queries to an input of the decoder and provides labels to only those anchors with ground-truth objects nearby. Following DAB-DETR and DN-DETR, DINO \cite{dino23} proposes a mixed object query selection method for anchor initialization and a look forward twice mechanism for box prediction. It provides the Contrastive DeNoising (CDN) module, which takes positional queries as anchor boxes and adds additional DN loss. In Figure~\ref{fig:Rego-Dino}, the bottom right block indicates DINO. This detector uses $\lambda_1$ and $\lambda_2$ hyperparameters where $\lambda_1<\lambda_2$. The bounding box $b=(x_i,y_i,w_i,h_i)$ taken as input in the decoder, its corresponding generated anchor is denoted as $a=(x_i,y_i,w_i,h_i)$.
\begin{align}
\resizebox{.87\hsize}{!}{$ ATD(k)=\frac{1}{k}\Sigma\{M_K(\{\parallel b_0-a_0\parallel_1,\parallel b_1-a_1\parallel_1,...,\parallel b_{N-1}-a_{N-1}\parallel_1\},k)\}$}
\end{align}
Where $\parallel(b_i-a_i)\parallel$ is the distance between the anchor and bounding box and $M_K(x, k)$ is the function that provides the top K elements in x. The $\lambda$ parameter is the threshold value for generating noise for anchors that are fed as input object queries to the decoder. It provides two types of anchor queries: positive with threshold value less than $\lambda_1$ and negative with noise threshold values greater than $\lambda_1$ and less than $\lambda_2$. This way, the anchors with no ground-truth nearby are labeled as "no object". Thus, DINO makes the training convergence faster and improves performance for small objects. 
\begin{figure*}[h]
\centering
\includegraphics[width=0.9\textwidth]{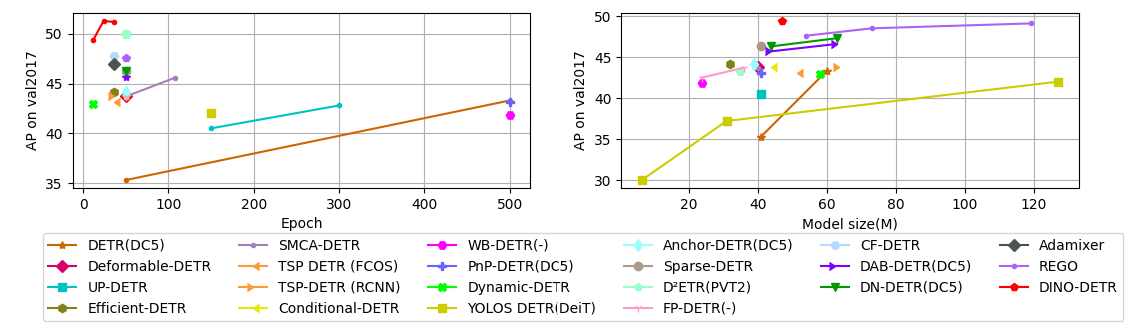}
\caption{Comparison of all DETR-based detection transformers on COCO minival set. (a) Performance comparison of detection transformers using a ResNet-50 \cite{resnet5} backbone w.r.t. training epochs. Networks that are labeled with DC5 take a dilated feature map. (b) Performance comparison of detection transformers w.r.t. model size (parameters in million).}\label{fig:Epoch-AP-size}
\end{figure*}
\begin{figure*}[h]
\centering
\includegraphics[width=0.9\textwidth]{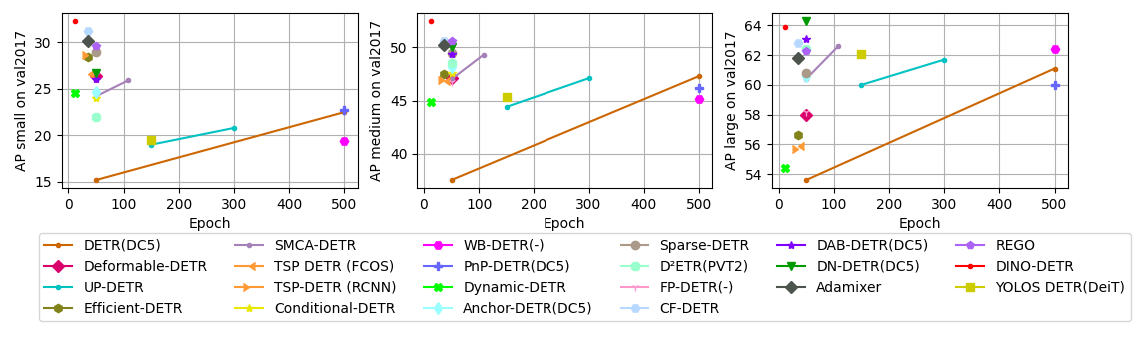}
\caption{Comparison of DETR-based detection transformers on COCO minival set using a ResNet-50 backbone. (a) Performance comparison of detection transformers on small objects. (b) Performance comparison of detection transformers on medium objects. (c) Performance comparison of detection transformers on large objects. }\label{fig:Epoch-AP-S-M-L}
\end{figure*}
\section{Datasets and Evaluation Metrics}
\label{sec:datasetEV}
It is important to compare modifications in detection transformers to understand their effect on network size, training convergence, and performance. This section presents quantitative comparisons of improvements in DETR on popular benchmark MS COCO \cite{coco14}. A mini val set of the COCO2014 is used for detection transformers evaluation. These results are evaluated using mean Average Precision (mAP) as the evaluation metric. The mAP is the mean of each object category's Average Precision (AP), where AP is the area under the precision-recall curve \cite{pr61}. 

\begin{table*}
\begin{center}
\caption{Performance comparison of all DETR-based detection transformers on COCO minival set. Here, networks labeled with DC5 take a dilated feature map. The IoU threshold values are set to 0.5 and 0.75 for AP calculation and also calculate AP for small( $AP_s$), medium ($AP_m$) and large ($AP_l$) objects. + represents bounding-box refinement and ++ denotes Deformable-DETR. $\ast$$\ast$ indicates Efficient-DETR used 6-encoder layers and 1-decoder layer. S denotes Small, and B indicates Base. \dag~ represents the distillation mechanism by Touvron et al. \cite{deit77}. \ddag~indicates the model is Pre-trained on ImageNet-21k. All models use 300 queries, while DETR uses 100 object queries to input to the decoder network. The models with superscript $\ast$ use three pattern embeddings. The three best results are represented in \textcolor{red}{red}, \textcolor{blue}{blue}, and \textcolor{green}{green}, respectively.}\label{tab:tablecomp}
\renewcommand{\arraystretch}{0.7} 
\begin{tabular*}{\textwidth}
{@{\extracolsep{\fill}}p{2.6cm}p{2.4cm}p{1.8cm}p{0.6cm}p{0.9cm}p{1.1cm}p{0.6cm}p{0.6cm}p{0.6cm}p{0.6cm}p{0.6cm}p{0.6cm}@{\extracolsep{\fill}}}
\toprule
\textbf{Methods} & 
\textbf{Backbone} &
\textbf{Publications} &
\textbf{Epoch} & 
\textbf{GFLOPs} & 
\textbf{Parameters (M)} & 
\textbf{AP} & 
\textbf{AP\textsuperscript{50}} &
\textbf{AP\textsuperscript{75}}  & \textbf{AP\textsubscript{S}} & 
\textbf{AP\textsubscript{M}} &
\textbf{AP\textsubscript{L}} \\
\toprule
& DC5-ResNet-50 & & 109 & 320 & 166 & 41.1 & 61.4 & 44.3 & 22.9 & 45.9 & 55.0 \\
\multirow{1}{*}{Faster R-CNN \cite{faster23}} & ResNet-50-FPN & CVPR 2015  & 109 & 180  & 42 & 42.0 & 62.1 & 45.5 & 26.6 & 45.4 & 53.4\\
& ResNet-101-FPN & & 109 & 246  & 60 & 44.0 & 63.9  & 47.8 & 27.2 & 48.1 & 56.0  \\
\midrule

& DC5-ResNet-50 & & 50 & 187  & 41 & 35.3 & 55.7 & 36.8 & 15.2 & 37.5 & 53.6 \\
\multirow{1}{*}{DETR \cite{detr34}} & DC5-ResNet-50 & ECCV 2020 & 500 & 187  & 41 & 43.3 & 63.1 & 45.9 & 22.5 & 47.3 & 61.1 \\
& DC5-ResNet-101 & & 500 & 253  & 60 & 44.9 & 64.7 & 47.7 & 23.7 & 49.5 & 62.3\\
\midrule

& ResNet-50 & & 50  & 173  & 40 & 43.8 & 62.6 & 47.7 & 26.4 & 47.1 & 58.0 \\
\multirow{1}{*}{Deformable-DETR \cite{Deformable54}} & ResNet-50+ & ICLR 2021 & 50  & 173  & 40 & 45.4 & 64.7 & 49.0  & 26.8 & 48.3 & 61.7  \\
& ResNet-50++  & & 50  & 173  & 40  & 46.2 & 65.2 & 50.0 & 28.8 & 49.2 & 61.7  \\
\midrule

\multirow{2}{*}{UP-DETR \cite{updetr23} } 
& ResNet-50 &  \multirow{2}{*}{CVPR 2021} & 150 & 86  & 41 & 40.5 & 60.8 & 42.6 & 19.0 & 44.4 & 60.0 \\
& ResNet-50 & & 300 & 86 & 41 & 42.8 & 63.0 & 45.3 & 20.8 & 47.1 & 61.7 \\
\midrule

& ResNet-50 && 36 & 159  & 32 & 44.2 & 62.2 & 48.0 & 28.4 & 47.5 & 56.6 \\
\multirow{1}{*}{Efficient-DETR \cite{efficientDE}} & ResNet-101 & arXiv 2021 & 36 & 239  & 51 & 45.2 & 63.7 & 48.8 & 28.8 & 49.1 & 59.0 \\
& ResNet-101 $\ast$ $\ast$ & & 36 & 289 & 54 & 45.7 & 64.1 & 49.5 & 28.2 & 49.1 & 60.2 \\
\midrule

& ResNet-50 & & 50 & 152  & 40 & 43.7 & 63.6 & 47.2 & 24.2 & 47.0 & 60.4 \\
\multirow{1}{*}{SMCA-DETR \cite{smca23}} & ResNet-50 & ICCV 2021 & 108 & 152  & 40 & 45.6 & 65.5 & 49.1 & 25.9 & 49.3 & 62.6 \\
& ResNet-101 & & 50 & 218  & 58 & 44.4 & 65.2 & 48.0 & 24.3 & 48.5 & 61.0  \\
\midrule

\multirow{2}{*}{TSP-DETR \cite{tspdetr81}} 
& FCOS-ResNet-50 & \multirow{2}{*}{ICCV 2021} & 36 & 189  & 51.5 & 43.1 & 62.3 & 47.0 & 26.6 & 46.8 & 55.9 \\
& RCNN-ResNet-50 && 36 & 188  & 63.6 & 43.8 & 63.3 & 48.3 & 28.6 & 46.9 & 55.7 \\
\midrule

\multirow{2}{*}{Conditional-DETR \cite{CondDE}} 
& DC5-ResNet-50 & \multirow{2}{*}{ICCV 2021} & 50 & 195  & 44 & 43.8 & 64.4 & 46.7 & 24.0 & 47.6 & 60.7\\
& DC5-ResNet-101 && 50 & 262  & 63 & 45.0 & 65.5 & 48.4 & 26.1 & 48.9 & 62.8 \\
\midrule

\multirow{1}{*}{WB-DETR \cite{WBdetr4}} & - & ICCV 2021 & 500 & 98  & 24 & 41.8 & 63.2 & 44.8 & 19.4 & 45.1 & 62.4\\
\midrule

\multirow{1}{*}{PnP-DETR \cite{pnp6}} 
& DC5-ResNet-50 & ICCV 2021 & 500  & 145 & 41 & 43.1 & 63.4 & 45.3 & 22.7 & 46.5 & 61.1 \\
\midrule

\multirow{1}{*}{Dynamc-DETR \cite{DynamicDE}} 
& ResNet-50 & ICCV 2021 & 12  & - & 58 & 42.9 & 61.0 & 46.3 & 24.6 & 44.9 & 54.4  \\
\midrule

\multirow{2}{*}{YOLOS-DETR \cite{yolos6}} & DeiT-S \cite{deit77} \dag &  \multirow{2}{*}{NeurIPS 2021}  & 150 & 194  & 31 & 36.1 & 56.5 & 37.1 & 15.3 & 38.5 & 56.2 \\
& DeiT-B \cite{deit77} \dag & & 150 & 538 & 127 & 42.0 & 62.2 & 44.5 & 19.5 & 45.3 & 62.1 \\
\midrule

\multirow{2}{*}{Anchor-DETR \cite{Anchor-detr}}
& DC5-ResNet-50 $\ast$ & \multirow{2}{*}{AAAI 2022} & 50 & 151 & 39 & 44.2 & 64.7 & 47.5 & 24.7 & 48.2 & 60.6  \\
& DC5-ResNet-101 $\ast$ & & 50 & 237 & 58 & 45.1 & 65.7 & 48.8 & 25.8 & 49.4 & 61.6 \\
\midrule

\multirow{2}{*}{Sparse-DETR \cite{sparsedetr}} 
& ResNet-50-$\rho$-0.5 & \multirow{2}{*}{ICLR 2022} & 50 & 136  & 41 & 46.3 & 66.0 & 50.1 & 29.0 & 49.5 & 60.8\\
& Swin-T-$\rho$-0.5 \cite{swins3} & & 50 & 144 & 41 & 49.3 & 69.5 & 53.3 & 32.0 & 52.7 & 64.9\\
\midrule

$D^2$ETR \cite{decoderonly60} & PVT2 & \multirow{2}{*}{arXiv 2022} & 50 & 82  & 35 & 43.2 & 62.9 & 46.2 & 22.0 & 48.5 & 62.4  \\
Def $D^2$ETR \cite{decoderonly60} & PVT2 & & 50 & 93 & 40 & 50.0 & 67.9 & 54.1 & 31.7 & 53.4 & \textcolor{blue}{66.7} \\
\midrule

FP-DETR-S \cite{fpdetr} & - && 50 & 102  & 24 & 42.5 & 62.6 & 45.9 & 25.3 & 45.5 & 56.9\\
FP-DETR-B \cite{fpdetr} & - & ICLR 2022 & 50 & 121 & 36 & 43.3 & 63.9 & 47.7 & 27.5 & 46.1 & 57.0\\
FP-DETR-B \ddag \cite{fpdetr} & - & &50 & 121 & 36 & 43.7 & 64.1 & 47.8 & 26.5 & 46.7 & 58.2\\
\midrule

\multirow{2}{*}{CF-DETR \cite{cfdetr}} 
& ResNet-50 & \multirow{2}{*}{AAAI 2022} & 36 & -  & - & 47.8 & 66.5 & 52.4 & 31.2 & 50.6 & 62.8\\
& ResNet-101 & & 36 & - & - & 49.0 & 68.1 & 53.4 & 31.4 & 52.2 & 64.3 \\
\midrule

\multirow{2}{*}{DAB-DETR \cite{dab89}} 
& DC5-ResNet-50 $\ast$ & \multirow{2}{*}{ICLR 2022} & 50 & 216 &  44 & 45.7 & 66.2 & 49.0 & 26.1 & 49.4 & 63.1  \\
& DC5-ResNet-101 $\ast$ & & 50 & 296 & 63 & 46.6 & 67.0 & 50.2 & 28.1 & 50.5 & 64.1\\
\midrule

\multirow{4}{*}{DN-DETR \cite{dn42}} 
& ResNet-50 & \multirow{4}{*}{CVPR 2022} &50 & 94 & 44 & 44.1 & 64.4 & 46.7 & 22.9 & 48.0 & 63.4\\
& DC5-ResNet-50 && 50 & 202  & 44 &46.3 & 66.4 & 49.7 & 26.7 & 50.0 & 64.3 \\
& ResNet-101 && 50 & 174  & 63 & 45.2 & 65.5 & 48.3 & 24.1 & 49.1 & 65.1 \\
& DC5-ResNet-101 && 50 & 282 & 63 & 47.3 & 67.5 & 50.8 & 28.6 & 51.5 & 65.0 \\
\midrule

\multirow{3}{*}{AdaMixer \cite{adamixer7}} 
& ResNet-50 & \multirow{3}{*}{CVPR 2022} &36 & 132 & 139 & 47.0 & 66.0 & 51.1 & 30.1 & 50.2 & 61.8\\
& ResNeXt-101-DCN & &36 & 214  & 160 & \textcolor{green}{49.5} & 68.9 & 53.9 & 31.3 & 52.3 & \textcolor{green}{66.3} \\
& Swin-s \cite{swins3}  & & 36 & 234  & 164 & \textcolor{red}{51.3} & \textcolor{red}{71.2} & \textcolor{green}{55.7} & \textcolor{green}{34.2} & \textcolor{red}{54.6} & \textcolor{red}{67.3} \\
\midrule

\multirow{3}{*}{REGO \cite{rego2}} 
&  ResNet-50++ & \multirow{3}{*}{CVPR 2022} & 50 & 190 & 54 & 47.6 & 66.8 & 51.6 & 29.6 & 50.6 & 62.3\\
&  ResNet-101++ && 50 & 257  & 73 & 48.5 & 67.0 & 52.4 & 29.5 & 52.0 & 64.4 \\
&  ReNeXt-101++ &&  50 & 434  & 119 & 49.1 & 67.5 & 53.1 & 30.0 & 52.6 & 65.0 \\
\midrule

\multirow{4}{*}{DINO\cite{dino23}} 
&  ReNet-50-4scale $\ast$ &  \multirow{4}{*}{arXiv 2022}& 12 & 279 & 47 & 49.0 & 66.6 & 53.5 & 32.0 & 52.3 & 63.0 \\
&  ResNet-50-5scale $\ast$ && 12 & 860 & 47 & 49.4 & 66.9 & 53.8 & 32.3 & 52.5 & 63.9 \\
&  ReNet-50-5scale $\ast$ && 24 & 860 &  47 & \textcolor{red}{51.3} & \textcolor{blue}{69.1} &\textcolor{red}{56.0} & \textcolor{blue}{34.5} & \textcolor{green}{54.2} & 65.8 \\
& ResNet-50-5scale $\ast$ && 36 & 860 & 47 & \textcolor{blue}{51.2} & \textcolor{green}{69.0} & \textcolor{blue}{55.8} & \textcolor{red}{35.0} & \textcolor{blue}{54.3} & 65.3 \\
\bottomrule
\end{tabular*}
\end{center}
\end{table*} 
\section{Results and Discussion}
\label{sec:comp}

Many advancements are proposed in DETR, such as backbone modification, Query design and attention refinement to improve performance and training convergence. Table~\ref{tab:tablecomp} shows the performance comparison of all DETR-based detection transformers on the COCO minival set. We can observe that DETR performs well at 500 training epochs and has low AP on small objects. The modified versions improve performance and training convergence like DINO has mAP of 49.0$\%$ at 12 epochs and performs well on small objects. 

\begin{table*}
\tiny
\begin{center}
\caption{Overview of Advantages and limitations of Detection Transformers.}\label{tab:adv-limit}
\renewcommand{\arraystretch}{0.6} 
\begin{tabular*}{\textwidth}
{@{\extracolsep{\fill}}p{1.9cm}p{1.1cm}p{6.5cm}p{6.5cm}@{\extracolsep{\fill}}}
\toprule
\textbf{Methods} & 
\textbf{Publications} &
\textbf{Advantages} & 
\textbf{Limitations} \\
\toprule
DETR \cite{detr34} & ECCV 2020 &  Removes the need for hand-designed components like NMS or anchor generation. & Low performance on small objects and slow training convergence.  \\
\midrule
Deformable-DETR \cite{Deformable54} &  ICLR 2021 &  Deformable attention network, which makes training convergence faster.  & Number of encoder tokens increases by 20 times compared to DETR.  \\
\midrule
UP-DETR \cite{updetr23} &   CVPR 2021 & Pre-training for Multi-tasks learning and Multi-queries localization. &  Pre-training for patch localization, CNN and transformers pre-training needs to integrate. \\
\midrule
Efficient-DETR \cite{efficientDE} &  arXiv 2021 & Reduces decoder layers by employing dense and sparse set based network & Increase in GFLOPs twice compared to original DETR. \\
\midrule
SMCA-DETR \cite{smca23} &  ICCV 2021 & Regression-aware mechanism to increase convergence speed &   Low performance in detecting small objects.\\
\midrule
TSP-DETR \cite{tspdetr81} &  ICCV 2021 & Deals with issues of Hungarian loss and the cross-attention mechanism of Transformer.  & Uses proposals in TSP-FCOS and feature points in TSP-RCNN as in CNN-based detectors.  \\
\midrule
Conditional-DETR\cite{CondDE} &  ICCV 2021 & Conditional queries remove dependency on content embeddings and ease the training.  & Performs better than DETR and deformable-DETR for stronger backbones. \\
\midrule
WB-DETR \cite{WBdetr4} &  ICCV 2021 & Pure transformer network without backbone. & Low performance on small objects. \\
\midrule
PnP-DETR \cite{pnp6} & ICCV 2021 & Sampling module provides foreground and a small quantity of background features.  & Breaks 2d spatial structure by taking foreground tokens and reducing background tokens. \\
\midrule
Dynamic-DETR \cite{DynamicDE} &  ICCV 2021 & Dynamic attention provides small feature resolution and improves training convergence.  & Still dependent on CNN networks as convolution-based encoder and an ROI-based decoder.  \\
\midrule
YOLOS-DETR \cite{yolos6} & NeurIPS 2021 &  Convert ViT pre-trained on ImageNet-1k dataset into Object detector. & Pre-trained ViT still needs improvements as it requires long training epochs.\\
\midrule
Anchor-DETR \cite{Anchor-detr} &  AAAI 2022 & Object
queries as anchor points that predict multiple objects at one position. & Consider queries as 2D anchor points which ignore object scale. \\
\midrule
Spare-DETR \cite{sparsedetr} &  ICLR 2022 & Improve performance by updating tokens referenced by the decoder. & Performance is strongly dependent on the backbone specifically for large objects.\\
\midrule
$D^2$ETR \cite{decoderonly60}  & arXiv 2022 & Decoder-only transformer network to reduce computational cost. & Decreases computation comlexity significantly but has low performance on small objects.  \\
\midrule
FP-DETR \cite{fpdetr} &  ICLR 2022 & Pre-Training of the encoder-only transformer. &  Low performance on large objects.\\
\midrule
CF-DETR \cite{cfdetr} & AAAI 2022 & Refine coarse features to improve localization accuracy of small objects. & Addition of three new modules increase network size.  \\
\midrule
DAB-DETR \cite{dab89} &  ICLR 2022 &  Anchor-boxes as queries, attention for different scale objects.  & Positional prior for only foreground objects.\\
\midrule
DN-DETR \cite{dn42} & CVPR 2022 & Denoising training for positional-prior for foreground and background regions.  & Denoising training by adding positive noise to object queries ignoring background regions. \\
\midrule
AdaMixer \cite{adamixer7} & CVPR 2022 & Faster Convergence, Improves the adaptability of query-based decoding mechanism. & Large number of parameters.\\
\midrule
REGO \cite{rego2} & CVPR 2022 & Attention mechanism gradually focus on foreground regions more accurately. &  Multi-stage RoI-based attention modeling increases the number of parameters.\\
\midrule
DINO \cite{dino23} & arXiv 2022 & impressive results on small and medium-sized datasets & Performance drops for large size objects \\
\bottomrule
\end{tabular*}
\end{center}
\end{table*} 
The quantitative analysis of DETR and its updated versions regarding training convergence and model size on the COCO minival set is performed. Part (a) of Figure~\ref{fig:Epoch-AP-size} shows the mAP of the detection transformers using a ResNet-50 backbone with training epochs. The original DETR, represented with a brown line, has low training convergence. It has an mAP value of 35.3$\%$ at 50 training epochs and 44.9 $\%$ at 500 training epochs. Here, DINO, represented with a red line, converges at low training epochs and gives the highest mAP on all epoch values. The attention mechanism in DETR involves computing pairwise attention scores between every pair of feature vectors, which can be computationally expensive, especially for large input images. Moreover, the self-attention mechanism in DETR relies on using fixed positional encodings to encode the spatial relationships between the different parts of the input image. This can slow down the training process and increase converging time. In contrast, Deformable-DETR and DINO have some modifications that can help speed up the training process. For example, Deformable DETR introduces deformable attention layers, which can better capture spatial context information and improve object detection accuracy. Similarly, DINO uses a denoising learning approach to train the network to learn more generalized features useful for object detection, making the training process faster and more effective.

Part (b) of Figure~\ref{fig:Epoch-AP-size} compares all detection transformers regarding the model size. Here, YOLOS-DETR uses DeiT-small as the backbone instead of DeiT-Ti, but it also increases model size by 20x times. DINO and REGO-DETR have comparable mAP, but REGO-DETR is nearly double in model size than DINO. These networks use more complex architectures than the original DETR architecture, which increase the total parameters and the overall network size.

We also provide a qualitative analysis of DETR and its updated versions on all-sized objects in Figure~\ref{fig:Epoch-AP-S-M-L}. For small objects, the mAP for the original DETR is 15.2$\%$ at 50 epochs, while Deformable-DETR has an mAP value of 26.4$\%$ at 50 epochs. The self-attention mechanism in Deformable-DETR allows it to interpolate features from neighboring pixels, which is particularly useful for small objects that may only occupy a few pixels in an image. This mechanism in Deformable-DETR captures more precise and detailed information about small objects, which can lead to better performance than DETR.

\section{Open Challenges \& Future Directions}
\label{sec:future}
Detection Transformers have shown promising results on various object detection benchmarks. There are still some open challenges and future directions for improving it. Table~\ref{tab:adv-limit} provides the advantages and limitations of all proposed improved versions of DETR.
Here are some open challenges and future directions for improvements in DETR:

\noindent\textbf{Improve attention mechanisms:}~The performance of detection transformers depends on the attention mechanism for capturing dependencies between various spatial locations in an image. Till now, 60\% of modifications have been done in the attention mechanism of the detection transformer to improve performance and training convergence. Future research could focus on designing more refined attention mechanisms to capture spatial information or incorporate task-specific constraints

\noindent\textbf{Adaptive and dynamic backbones:}~Backbone also affects the network performance and size. Current detection transformers remove the backbone or use fixed backbone architectures across all images. Only 10\% of backbone modifications are done in DETR to improve performance and reduce network size. Future research could explore dynamic backbone architectures that can adjust their complexity based on the input image's characteristics. Researchers can improve detection transformers by modifying the backbone, likely leading to even more impressive results.

\noindent\textbf{Improve quantity and quality of object queries:}~The quantity object queries fed to the decoder as input in DETR is typically fixed during training and inference. However, the size or number of objects in an image can differ. Later on, it is observed in some networks such as DAB-DETR, DN-DETR and DINO that modifying the quantity or quality of object queries can significantly affect the detection transformer performance. DAB-DETR uses dynamic anchor boxes as object queries, DN-DETR adds positive noise to the object queries for denoising training, and DINO adds positive and negative noise to the object queries for improved denoising training. Future models can adjust the number of object queries based on the image's content to improve the quantity of object queries. Furthermore, researchers can include more dynamic and adaptive mechanisms to improve the quality of object queries.

\section{Conclusion}
\label{sec:conclusion}
Detection transformers have provided efficient and precise object detection networks and delivered insights into the operation of deep neural networks. This review gives a detailed overview of the Detection Transformers. Specifically, it focuses on the latest advancements in DETR to improve performance and training convergence. The attention module of the detection transformer in the encoder-decoder network is modified to improve training convergence, and object queries as input to the decoder are updated to enhance the performance of small objects. We provide the latest improvements in detection transformers, including backbone modification, query design, and attention refinement. We also compare the advantages and limitations of detection transformers in terms of performance and architectural design. With its focus on object detection tasks, this review provides a unique view of the recent advancement in DETR. We hope this study will increase the researcher's interest in solving existing challenges towards applying transformers models in the object detection domain.


%
\ifCLASSOPTIONcompsoc
  \section*{Acknowledgments}
\else
  \section*{Acknowledgment}
\fi
 The work has been partially funded by the European project AIRISE under Grant Agreement ID 101092312.
\ifCLASSOPTIONcaptionsoff
  \newpage
\fi



\bibliographystyle{IEEEtran}
\bibliography{sn-bibliography}
%


%

\begin{IEEEbiography}[{\includegraphics[width=1in,height=1.25in,clip,keepaspectratio]{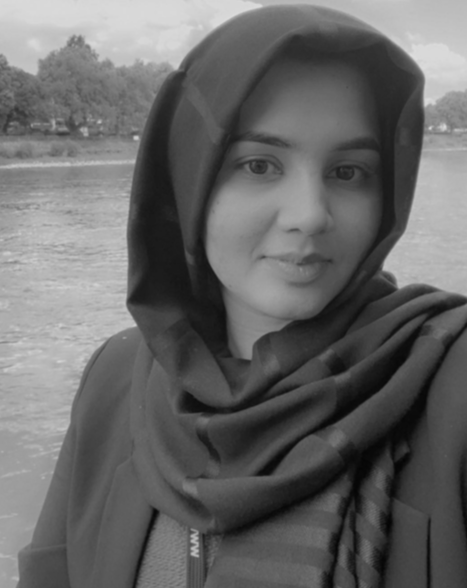}}]{TAHIRA SHEHZADI} received her bachelor's degree in electrical engineering from the University of Engineering and Technology Lahore, Pakistan and her M.S. degree in computer science from Pakistan Institute of Engineering and Applied Sciences, Pakistan. She is pursuing a PhD with the German Research Center for Artificial Intelligence (DFKI GmbH) and the Rheinland-Pf{\"a}lzische Technische Universität Kaiserslautern-Landau under the supervision of Prof . Didier Stricker and Dr. Muhammad Zeshan Afzal. Her research interests include deep learning for computer vision, specifically in 3D reconstruction. She received two Gold Medals for the Best Student from FAZAIA, Pakistan, in 2014 and secured University Merit Scholarship for a Master's degree in 2018 and the DAAD (Germany) PhD Fellowship in 2021.
\end{IEEEbiography}

\begin{IEEEbiography}[{\includegraphics[width=1in,height=1.25in,clip,keepaspectratio]{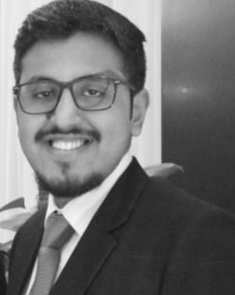}}]{KHURRAM AZEEM HASHMI} received his bachelor’s degree in Computer Science from the National University of Computer and Emerging Sciences, Pakistan, in 2016, and his M.S. degree from the Technical University of Kaiserslautern. He is currently a researcher at the German Research Center for Artificial Intelligence (DFKI) and pursuing a Ph.D. degree from RPTU Kaiserslautern-Landau under the supervision of Prof. Didier Stricker and Dr. Muhammad Zeshan Afzal. His research interests include self-supervised learning and instance-based representation learning in challenging conditions, such as in videos and dark environments. Alongside his research, he serves as a reviewer for major computer vision conferences and regularly reviews articles for journals, including IEEE Access, Springer Nature, Sensors, and Neurocomputing.
\end{IEEEbiography}

\begin{IEEEbiography}[{\includegraphics[width=1in,height=1.25in,clip,keepaspectratio]{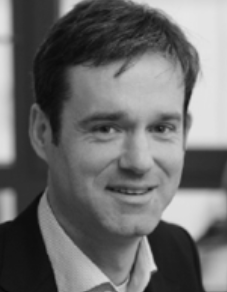}}]{DIDIER STRICKER} lead the Department of Virtual and Augmented Reality, Fraunhofer Institute for Computer Graphics (Fraunhofer IGD), Darmstadt, Germany, from June 2002 to June 2008. In this function, he initiated and participated in many national and international projects in the areas of computer vision and virtual and aug- mented reality. He is currently a Professor with the University of Kaiserslautern and the Scientific Director of the German Research Center for Artificial Intelligence (DFKI), Kaiserslautern, where he leads the Research Department of Augmented Vision. In 2006, he received the Innovation Prize from the German Society of Computer Science. He serves as a reviewer for different European or national research organizations, and is a regular reviewer of the most important journals and conferences in the areas of VR/AR and computer vision.
\end{IEEEbiography}

\begin{IEEEbiography}[{\includegraphics[width=1in,height=1.25in,clip,keepaspectratio]{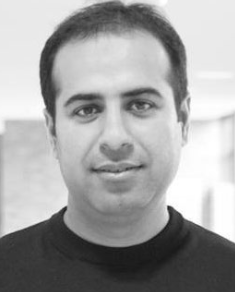}}]{MUHAMMAD ZESHAN AFZAL} received the master’s degree majoring in visual computing from Saarland University, Germany, in 2010, and the Ph.D. degree majoring in artificial intelligence from the Kaiserslautern University of Technology, Kaiserslautern, Germany, in 2016. He worked both in the industry (Deep Learning and AI Lead Insiders Technologies GmbH) and academia (TU Kaiserslautern). At an application level, his experience includes a generic segmentation framework for natural, human activity recognition, document and medical image analysis, scene text detection, recognition, and online and offline gesture recognition. Moreover, a special interest in recurrent neural networks and transformers for sequence processing applied to images and videos. He also worked with numerics for tensor-valued images. His research interests include deep learning for vision and language understanding. He is a member of IAPR. He received the Gold Medal for the Best Graduating Student in computer science from IUB, Pakistan, in 2002 and secured the DAAD (Germany) Fellowship in 2007.
\end{IEEEbiography}







\end{document}